\documentclass[runningheads]{llncs}
\usepackage[T1]{fontenc}
\usepackage{cite}
\usepackage{amsmath,amssymb,amsfonts}
\usepackage{colortbl}
\usepackage{xcolor}
\usepackage{subcaption}
\usepackage{makecell}
\usepackage{tcolorbox}
\usepackage{amsfonts}
\usepackage{multirow}
\usepackage{diagbox}
\usepackage{wrapfig}
\usepackage{caption}
\usepackage{hyperref}
\begin{document}
\title{GenFT: A Generative Parameter-Efficient Fine-Tuning Method for Pretrained Foundation Models}
\titlerunning{GenFT: A Generative PEFT Method}
%
\author{Guangning Xu\inst{1} \and
Baoquan Zhang\inst{2}\thanks{Corresponding author.} \and
Michael K. Ng \inst{1}}
\institute{Department of Mathematics, Hong Kong Baptist University, Hong Kong, China
\email{\{xuguangning,michael-ng\}@hkbu.edu.hk}\\
\and
Department of Computer Science, Harbin Institute of Technology, Shenzhen, China\\
\email{baoquanzhang@hit.edu.cn}
}

\maketitle              
\begin{abstract}
Parameter-efficient fine-tuning (PEFT) has emerged as a resource-efficient strategy for adapting Pretrained Foundation Models (PFMs) by learning a small number of task-specific updates $\Delta W$. Existing methods often learn $\Delta W$ largely independently of pretrained weights $W_0$, or exploit $W_0$ mainly through initialization or simple reparameterization. To further leverage the structural information encoded in $W_0$, we propose Generative Parameter-Efficient Fine-Tuning (GenFT), a $W_0$-conditioned PEFT method that uses a deterministic weight generator to produce task-specific updates. Specifically, GenFT performs row and column transformations with nonlinear activations to extract structured patterns from $W_0$, and introduces a shared-specific decomposition to balance cross-layer information reuse and layer-specific flexibility. GenFT is simple and parameter-efficient, achieving competitive or better average performance across NLP and CV benchmarks. We further provide a pilot study on LLaMA-7B to examine its feasibility for generative models. The code is available at GitHub \href{https://github.com/xuguangning1218/GenFT}{https://github.com/xuguangning1218/GenFT}.

\end{abstract}
\section{Introduction}
\label{introduction} 

PEFT is a fundamental strategy for adapting PFMs, which updates only a small fraction of parameters in large models. Among PEFT approaches, LoRA is arguably the most prominent: it employs a low-rank decomposition to parameterize weight updates, typically written as $\Delta W = AB$. Many PEFT methods can be viewed as learning task-specific updates under a prior over $\Delta W$, which can be written as
\begin{equation}
\label{eq:old_assumption}
\Delta W^* := \arg\max_{\Delta W \in \mathcal{S}} P(\mathcal{D} \mid W_0 + \Delta W) P(\Delta W).
\end{equation}
This formulation treats $\Delta W$ as the main trainable object, while the pretrained weights $W_0$ mainly serve as the fixed backbone to which $\Delta W$ is added. Recently, several PEFT methods have started to explicitly exploit $W_0$. For example, PiSSA~\cite{meng2024pissa} uses the SVD of $W_0$ to initialize adapters, while DoRA~\cite{liu2024dora} decomposes weights into magnitude and direction. However, these methods usually exploit $W_0$ through initialization, decomposition, or element-wise interaction~\cite{huang2025hira}, rather than using $W_0$ as a direct condition for generating task-specific updates throughout optimization. Motivated by this observation, we revisit PEFT from a conditional weight-generation perspective. Instead of learning $\Delta W$ largely independently of $W_0$, we consider a conditional prior over updates:
\begin{equation}
\label{eq:assumption}
\Delta W^* :=\arg\max_{\Delta W \in \mathcal{S}} P(\mathcal{D} \mid W_0 + \Delta W) P(\Delta W \mid W_0)P(W_0).
\end{equation}
Here, Eq.~\eqref{eq:assumption} is used as a conceptual motivation rather than a full probabilistic generative model. In practice, we instantiate this conditional perspective with a deterministic generator $G_{\theta}(W_0)$, which produces task-specific updates by extracting structured information from the pretrained weights.

Finally, we introduce Generative Parameter-Efficient Fine-Tuning (GenFT), a $W_0$-conditioned PEFT method that uses a deterministic generator to produce task-specific updates $\Delta W$. Specifically, GenFT applies row and column transformations with activation functions to extract structured information from $W_0$ for update generation. Moreover, we introduce a shared-specific decomposition of the transformation factors, enabling information reuse across layers while preserving layer-specific flexibility. Our contributions are summarized as follows:
\begin{itemize}
\item We propose GenFT, a $W_0$-conditioned PEFT framework that generates task-specific updates through row and column transformations, improving adaptation across NLP and CV tasks.
\item We introduce a shared-specific decomposition of the transformation factors, capturing cross-layer shared patterns and layer-specific variations with a small parameter budget.
\item Experiments on VTAB-1K with ViT and GLUE with RoBERTa$_{\textit{Base}}$ show that GenFT achieves competitive or better average performance than representative PEFT baselines, and a pilot study on Alpaca with LLaMA-7B further examines its feasibility for generative models.
\end{itemize}

\section{Related Work}
\label{related_work}

\textbf{Additive PEFT} methods introduce lightweight layers into transformers to enhance adaptability for customized tasks. Adapter~\cite{houlsby2019parameter} inserts a downsampling-activation-upsampling layer after the feedforward module. AdaptFormer~\cite{chen2022adaptformer} targets the MLP layer, improving transferability without modifying its parameters. AdapterDrop~\cite{ruckle2021adapterdrop} dynamically disables adapters in lower layers during training to eliminate ineffective modules and boost efficiency. UniPT~\cite{diao2024unipt} accelerates training by adding parallel learnable modules to adapters, enabling simultaneous updates across layers. Tiny-Attention Adapter~\cite{zhao2022tiny} designs compact adapters with minimal per-head dimensionality, directly modifying latent representations based on cross-position dependencies. These methods improve adaptation by modifying activations or inserting additional modules, but they do not explicitly generate weight updates from the structure of pretrained weights.

\textbf{Selective PEFT} methods are designed to update a subset of parameters in a pretrained model based on their importance or task relevance. BitFit~\cite{zaken2022bitfit} tunes only the bias terms of each layer, achieving promising results on small-to-medium-scale datasets. SPT~\cite{he2023sensitivity} identifies sensitive parameters using gradient-based criteria, applying unstructured tuning to these parameters or switching to LoRA if the criteria are not met. GPS~\cite{zhang2024gradient} employs a gradient-based method to select and tune a small subset of sensitive parameters while keeping the remaining parameters frozen. LoRAPrune~\cite{zhang2024loraprune} uses LoRA-guided pruning to estimate structural importance, iteratively removing redundant channels and heads. SH-PEFT~\cite{liu2024sparsity} adopts a task-specific approach, selecting weights sensitive to the target task while discarding those responsive to unrelated tasks. These methods focus on identifying which parameters should be updated, while the generation of structured task-specific updates from $W_0$ remains less explored.

\textbf{Reparameterized PEFT} methods restructure model parameters using low-rank or low-dimensional representations. LoRA~\cite{hu2022lora}, a prominent method, reparameterizes updates to pretrained weights $W_0$ as $W_0^{\prime} = W_0 + AB$, where $A$ and $B$ are low-rank matrices. DoRA~\cite{liu2024dora} decomposes $W_0$ into magnitude and direction components to enhance parameter learning. HiRA~\cite{huang2025hira} applies the Hadamard product to combine LoRA weights with $W_0$, aiming to improve expressiveness but potentially disrupting neuron correlations. FourierFT~\cite{gaoparameter} transforms reparameterized weights into the Fourier domain, learning a small set of spectral coefficients. VB-LoRA~\cite{citation-0} introduces a shared vector bank with a differentiable top-$k$ module, enabling parameter sharing across layers. These methods exploit low-dimensional parameterization, initialization, decomposition, or element-wise interaction, but they do not explicitly model row- and column-wise transformations of $W_0$ for update generation.

Overall, existing PEFT methods mainly improve efficiency through aforementioned PEFT methods. In contrast, GenFT is complementary to these directions: it uses $W_0$ as the input to a deterministic update generator and extracts structured row- and column-wise information for task-specific adaptation.

\section{GenFT}
\label{methodology}

\begin{figure*}[htbp]
\centering
\includegraphics[width=0.9\linewidth]{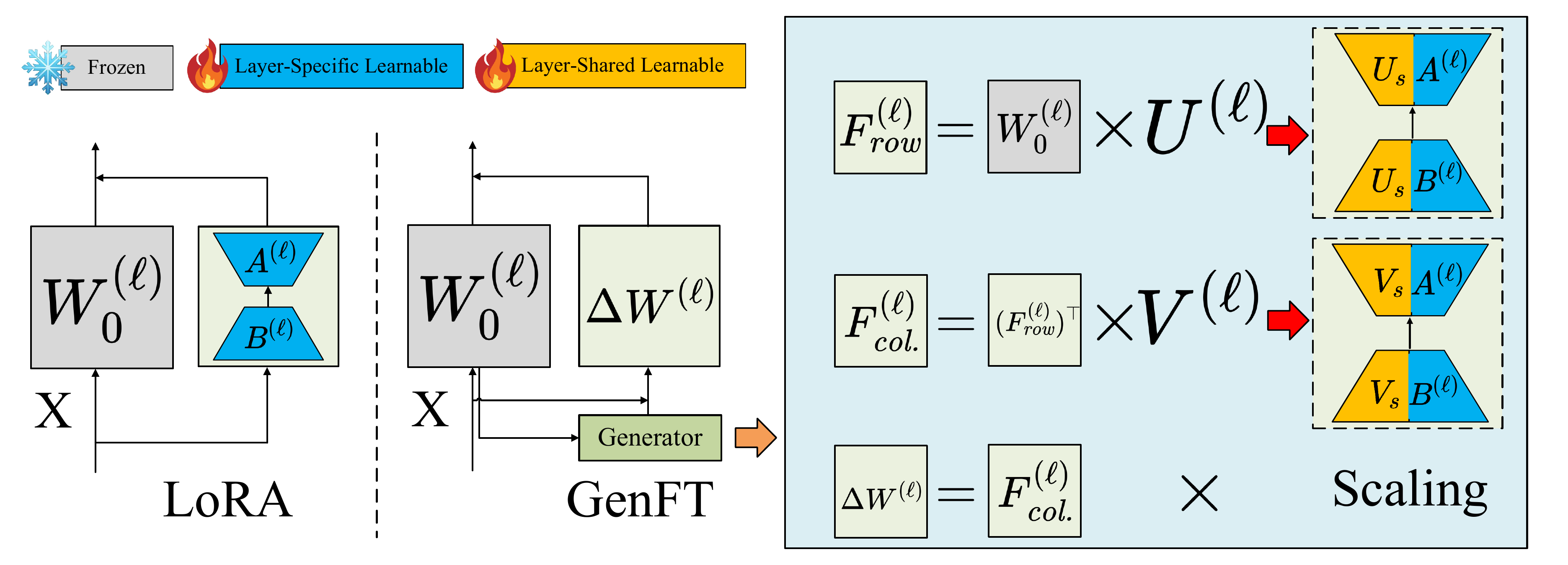}
\caption{Comparison of LoRA and GenFT. LoRA (left) learns task-specific updates under Eq.~\eqref{eq:old_assumption}. GenFT (middle) follows the conditional motivation in Eq.~\eqref{eq:assumption}. The GenFT generator (right) produces $\Delta W^{(\ell)} = \mathcal{G}_{\theta}(W^{(\ell)}_0)$ by extracting row- and column-wise features $F^{(\ell)}_{\textit{row}}$ and $F^{(\ell)}_{\textit{col.}}$ from $W^{(\ell)}_0$. The shared-specific decomposition balances cross-layer information reuse and layer-specific flexibility through latent transformation dimensions. Here, col. denotes column; activations are omitted for brevity.}
\label{fig:model}
\end{figure*}

Building on Eq.~\eqref{eq:assumption}, the GenFT is illustrated in Fig.~\ref{fig:model}. Instead of learning $\Delta W^{(\ell)}$ at $\ell$ layer independently of $W^{(\ell)}_0$, GenFT uses a deterministic generator to extract structured information from $W^{(\ell)}_0$ for update generation.

\textbf{Paradigm.} We assume that pretrained weights $W^{(\ell)}_0$ encode transferable structural information in their row and column spaces. GenFT therefore conditions the task-specific update $\Delta W^{(\ell)}$ on $W^{(\ell)}_0$ through a learnable generator. The core paradigm is:
\begin{equation}
h^{(\ell)} = W^{(\ell)}_0 X^{(\ell)} + \Delta W^{(\ell)} X^{(\ell)} = W^{(\ell)}_0 X^{(\ell)} + \mathcal{G}_{\theta}(W^{(\ell)}_0) X^{(\ell)},
\end{equation}
where $\mathcal{G}_{\theta}(\cdot)$ produces $\Delta W^{(\ell)} \in \mathbb{R}^{D\times D}$ from $W^{(\ell)}_0 \in \mathbb{R}^{D\times D}$. For clarity, we present the square-weight case; the formulation can be directly extended to non-square linear layers by matching input and output dimensions.

\textbf{Generator.} Since rows and columns of $W^{(\ell)}_0$ correspond to output-channel and input-feature structures, respectively, GenFT applies row and column transformations to extract complementary information from $W^{(\ell)}_0$. These transformations are controlled by learnable factors in $\theta$, and the resulting features are used to generate the task-specific update $\Delta W^{(\ell)}$.

\textbf{Row Transformation.} To balance shared and layer-specific information, we decompose the row-transformation matrix into a layer-shared component and a layer-specific component. The shared factors are reused across layers, while the layer-specific factors are learned independently for each layer. For the $\ell$-th layer, the row-transformation matrix is formulated as:
\begin{equation}
\begin{aligned}
U_1^{(\ell)} &= \begin{pmatrix}
U_s & A^{(\ell)}
\end{pmatrix},\quad
U_2^{(\ell)} = \begin{pmatrix}
U_s & B^{(\ell)}
\end{pmatrix}, \\
U^{(\ell)} &= U_2^{(\ell)} (U_1^{(\ell)})^\top
= U_s U_s^\top + B^{(\ell)} (A^{(\ell)})^\top,
\end{aligned}
\end{equation}
where $U_s \in \mathbb{R}^{D \times a}$ captures layer-shared row-wise information, and $A^{(\ell)}, B^{(\ell)} \in \mathbb{R}^{D \times b}$ capture layer-specific information. The row transformation is:
\begin{equation}
F^{(\ell)}_{\textit{row}} = \sigma_1(\textit{ratio} \cdot W_0 U^{(\ell)})\odot M_p,
\end{equation}
where $\sigma_1$ is an activation function, \textit{ratio} controls the proportion of $W^{(\ell)}_0$ used for update generation, and $M_p$ is a binary mask matrix that disables $p\%$ of the corresponding weights.

\textbf{Column Transformation.} Complementary to the row transformation, the column transformation extracts input-feature information from $W^{(\ell)}_0$. We use distinct layer-specific factors for the column transformation:
\begin{equation}
\begin{aligned}
V_1^{(\ell)} &= \begin{pmatrix}
V_s & A^{(\ell)}
\end{pmatrix},\quad
V_2^{(\ell)} = \begin{pmatrix}
V_s & B^{(\ell)}
\end{pmatrix}, \\
V^{(\ell)} &= V_2^{(\ell)} (V_1^{(\ell)})^\top
= V_s V_s^\top + B^{(\ell)} (A^{(\ell)})^\top,
\end{aligned}
\end{equation}
where $V_s \in \mathbb{R}^{D \times a}$ captures layer-shared column-wise information, and $A^{(\ell)}, B^{(\ell)}$ is shared parameters from row transformation. The update is obtained by:
\begin{equation}
\begin{aligned}
F^{(\ell)}_{\textit{col.}} &= \sigma_2\left(\left(F^{(\ell)}_{\textit{row}}\right)^{\top} V^{(\ell)}\right) \odot M_p,\
\Delta W^{(\ell)} &= \textit{scaling} \cdot F^{(\ell)}_{\textit{col.}},
\end{aligned}
\end{equation}
where $\sigma_2$ is an activation function and \textit{scaling} controls the magnitude of $\Delta W^{(\ell)}$. The latent dimension $a+b$ controls the transformation space of GenFT, but it should not be interpreted as the exact algebraic rank of the final $\Delta W^{(\ell)}$ after nonlinear activations and masking.

\begin{wrapfigure}{r}{0.50\textwidth}
    \centering
    \includegraphics[width=0.48\textwidth]{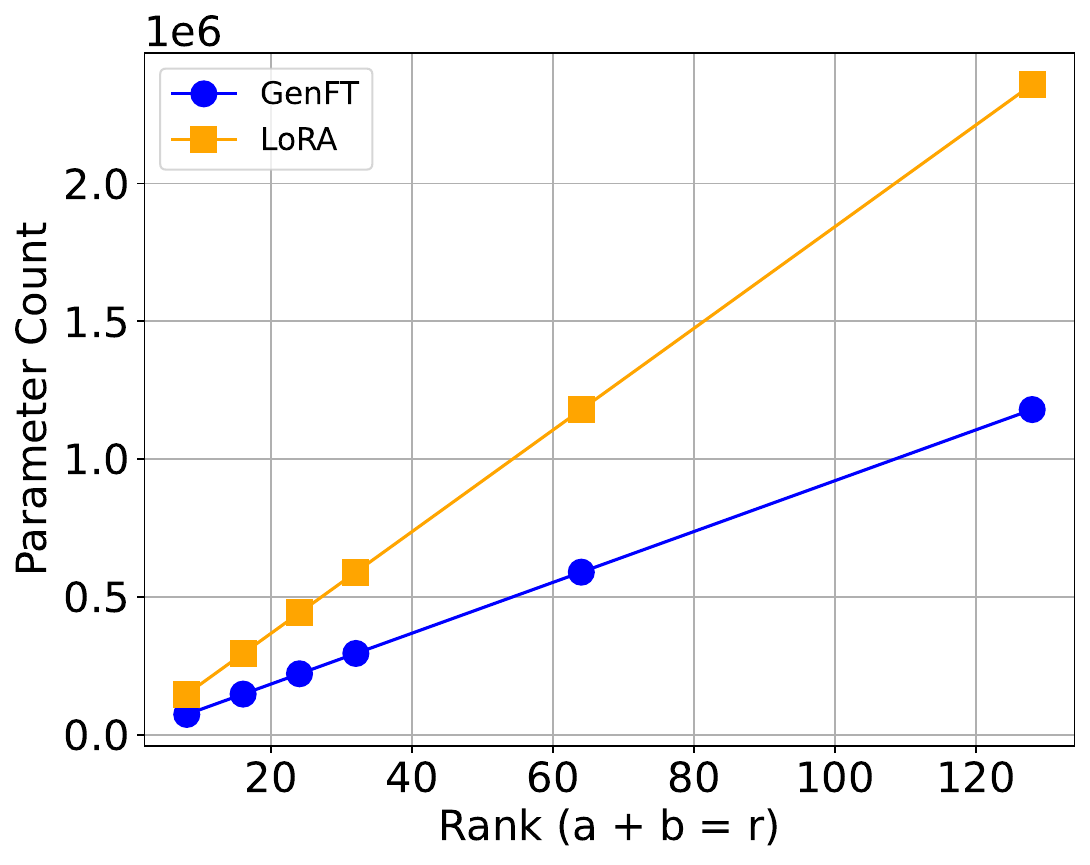}
    \caption{\# Params. vs Rank ($L=12$)}
    \label{fig:param_rank}
\end{wrapfigure}

\textbf{Finding.} Given a model with $L>1$ layers, let $a$ denote the layer-shared dimension, $b$ the layer-specific dimension, and $r$ denote LoRA's low-rank dimension. Under a comparable parameter budget, GenFT can use a larger latent transformation dimension than LoRA. See Appendix section A for more explanation.

As shown in Fig.~\ref{fig:param_rank}, LoRA’s parameter count grows rapidly with $r$, whereas GenFT grows more gradually with $a+b$. Here, $a+b$ denotes the latent transformation dimension rather than the exact algebraic rank of the final $\Delta W^{(\ell)}$, since nonlinear activations and masking may change the resulting matrix rank.

\begin{table*}[htbp]
\centering
\caption{Experimental results on GLUE benchmark~\cite{wangglue} with  RoBERTa$_{\textit{Base}}$~\cite{liu2019roberta}.}
\label{tab:glue}
\renewcommand\arraystretch{0.8}
\resizebox{\linewidth}{!}{
\begin{tabular}{c|cccccccccc}
\hline
\rowcolor{gray!25}\textbf{Method} & \makecell{\textbf{Param.} \\\textbf{(M)}} & \makecell{\textbf{CoLA} \\ \textbf{(Mcc.)}} & \makecell{\textbf{SST-2} \\\textbf{(Acc.)}} & \makecell{\textbf{MRPC} \\ \textbf{(Acc.)}}  & \makecell{\textbf{STS-B} \\ \textbf{(Pea.)}} & \makecell{\textbf{QQP} \\ \textbf{(Acc.)}} & \makecell{\textbf{MNLI} \\ \textbf{(Acc.)}} & \makecell{\textbf{QNLI} \\ \textbf{(Acc.)}} & \makecell{\textbf{RTE} \\ \textbf{(Acc.)}}  & \textbf{Avg.} \\
\hline
Full FT~\cite{kowsher2025propulsion} & 124.6 & 59.84 & 92.89 & 85.24 & 90.48 & 90.18 & 86.27 & 91.17 & 72.43 & 83.56 \\
Adapter$^S$~\cite{houlsby2019parameter} & 7.41 & 60.32 & 92.14 & 89.24 & 90.25 & 90.81 & \underline{87.33} & 90.84 & 73.56 & 84.31 \\
PromptTuning~\cite{lester2021power} & 0.61 & 49.37 & 91.09 & 74.83 & 82.44 & 82.99 & 80.57 & 80.03 & 58.12 & 74.93 \\
PrefixTuning~\cite{li2021prefix} & 0.96 & 55.31 & 92.17 & 87.25 & 88.48 & 87.75 & 85.21 & 90.77 & 54.51 & 80.18 \\
(IA)$^3$~\cite{liu2022few} & 0.66 & 59.58 & 92.02 & 87.00 & 90.30 & 87.99 & 83.95 & 90.88 & 71.12 & 82.85 \\
BitFit~\cite{zaken2022bitfit} & 0.086& 61.38 & 92.67 & 88.22 & 90.34 & 88.12 & 84.64 & 91.09 & 75.58 & 84.20 \\
LoRA~\cite{hu2022lora} & 0.89 & 60.09 & 92.40 & 88.50 & 90.66 & 88.83 & 86.54 & 92.02 & 72.92 & 83.99 \\
AdaLoRA~\cite{zhang2023adaptive} & 1.03 & 59.82 & 91.69 & 88.99 & 90.83 & 88.58 & 86.26 & 91.43 & 70.04 & 83.45 \\
MAM$_{\textit{Adapter}}$~\cite{hetowards} & 46.78 & 58.42 & 93.19 & 89.31 & 90.74 & 88.31 & 86.63 & 90.19 & 72.62 & 83.67 \\
PROPETL$_{v1}$~\cite{zeng2023one} & 1.87 & \underline{63.11} & 92.18 & 85.25 & \underline{91.33} & 89.22 & 86.49 & \underline{92.56} & 75.54 & 84.46 \\
PROPETL$_{v2}$~\cite{zeng2023one}  & 10.49 & 60.18 & 91.36 & 86.73 & 90.30 & 88.54 & 86.22 & 91.51 & 63.31 & 82.26 \\
PROPETL$_{v3}$~\cite{zeng2023one} & 1.77 & 61.72 & 92.54 & 87.42 & 90.76 & 88.90 & 86.84 & 92.06 & 67.39 & 83.45 \\
CorDA~\cite{yang2024corda} & 0.29 & 59.60 & 93.12 & \underline{89.71} & 90.17 & \textbf{91.10} & 87.28 & 91.49 & 76.17 & 84.83\\
Propulsion~\cite{kowsher2025propulsion} & 0.09 & 61.76 & 93.18 & 89.34 & \textbf{91.37} & 89.11 & 86.41 & \textbf{92.79} & 75.66 & 84.95 \\
GaLore~\cite{zhaogalore}	& -	& 60.35	& 94.04	& 87.01	& 90.73	& \underline{91.06}	& 87.00	& 92.24	& 79.42	& 85.23\\
FiRA~\cite{chen2024fira} & - & 59.83 &	93.58	& 88.97	& 90.33	& 90.30	& 86.25	& 91.69	& 75.09	& 84.50 \\
PiSSA~\cite{meng2024pissa}	& 0.30	& \textbf{66.03}	& 93.92	& 88.73	& 90.40	& 88.37	& 85.41	& 91.80	& \textbf{78.70}	& \underline{85.42}\\
LISA~\cite{pan2024lisa} & 53.17 & 62.29	& \underline{94.04}	& 89.22	& 89.72	& 88.37	& 84.79	& 90.72	& 77.26	& 84.55\\
\hline
\rowcolor{gray!25}GenFT & 0.24 & \textbf{63.38} & \textbf{94.84} & \textbf{89.95} & 90.69 & 90.59 & \textbf{87.38} & 92.49 & \underline{77.62} & \textbf{85.87} \\
\hline
\end{tabular}
}
\end{table*}

\section{Experiments}

\textbf{Natural Language Understanding}. We fine-tune the RoBERTa$_{\textit{Base}}$~\cite{liu2019roberta} with GenFT on GLUE benchmark~\cite{wangglue}. Our experimental setup follows~\cite{kowsher2025propulsion}. We perform random hyperparameter search over the layer-shared dimension $a$, layer-specific dimension $b$, activation functions $\sigma_1$ and $\sigma_2$, initialization schemes for $A^{(\ell)}$ and $B^{(\ell)}$, the $W^{(\ell)}_0$ ratio, and the scaling factor for $\Delta W^{(\ell)}$, using the AdamW optimizer with a preset weight decay rate. To ensure reproducibility, we fix all random seeds at 42.

\textbf{Results on GLUE.} Results are reported in Table~\ref{tab:glue}. GenFT ranks fourth on STS-B (behind PROPETL$_{\textit{LoRA}}$) and third on QQP and QNLI (behind Adapter$^S$ and PROPETL$_{v1}$), likely due to the sensitivity of these tasks to hyperparameter configurations. Nevertheless, GenFT achieves the best average score (85.87\%) with only 0.24M parameters, outperforming LoRA and other baselines on the remaining tasks. This improvement stems from its generative PEFT design: row--column transformations and the rank-decomposition policy, together with $\sigma_1$ and $\sigma_2$, extract structured information from $W_0^{(\ell)}$ into $\Delta W^{(\ell)}$, leading to robust performance across diverse NLU tasks.

\noindent\textbf{Image Classification}. We adopt the ViT-Base model~\cite{dosovitskiy2020image} (85.8M parameters), for multi-class and fine-grained image classification. We evaluate GenFT on VTAB-1K~\cite{zhai2019large}, which contains 19 sub-datasets spanning natural, specialized, and structured images. We also validate GenFT on FGVC, please refer to Appendix section D for more details. We report top-1 accuracy as the primary evaluation metric. Our experimental setup follows V-PETL Bench~\cite{xin2024v}. We conduct random hyperparameter search over the layer-shared dimension $a$, layer-specific dimension $b$, activation functions $\sigma_1$ and $\sigma_2$, and initialization schemes for $A^{(\ell)}$ and $B^{(\ell)}$, using the AdamW optimizer with a preset weight decay rate. To ensure reproducibility, we fix all random seeds.

\textbf{Results on VTAB-1K.} Results are reported in Table~\ref{tab:vtab}. GenFT achieves the highest average accuracy, outperforming all baselines on natural and specialized images and remaining competitive on structured datasets. Compared to full fine-tuning, GenFT avoids excessive computational costs. Moreover, GenFT’s row and column transformations with activation functions $\sigma_1$ and $\sigma_2$ enable flexible adaptation, surpassing additive PEFT methods such as AdaptFormer, while its rank-decomposition policy broadens the adaptation scope relative to selective PEFT. Compared to LoRA and its variants, GenFT further leverages $W^{(\ell)}_0$ via row and column transformations for robust $\Delta W^{(\ell)}$ generation.

\begin{table}[htbp]
\centering
\caption{Experimental results on VTAB-1K~\cite{zhai2019large} with ViT-B/16~\cite{dosovitskiy2020image} model. Details are shown in Appendix section H.}
\label{tab:vtab}
\renewcommand\arraystretch{0.8} 
\resizebox{\linewidth}{!}{
\begin{tabular}{c|ccccc}
\hline
\rowcolor{gray!25}\textbf{Method} & \makecell{\textbf{Param.(M)}} & \makecell{\textbf{Nat.(Acc.)}} & \makecell{\textbf{Spe.(Acc.)}} & \makecell{\textbf{Str.(Acc.)}} & \textbf{Avg.} \\ 
\hline
Full FT~\cite{jia2022visual} & 85.88 & 68.74 & 83.38 & 47.64 & 65.57 \\
Linear~\cite{jia2022visual} & 0.00 & 68.93 & 77.15 & 26.85 & 52.94 \\ 
Adapter~\cite{houlsby2019parameter} & 0.16 & 79.01 & 84.08 & 58.49 & 71.44 \\
VPT~\cite{jia2022visual} & 0.56 & 78.49 & 82.42 & 55.00 & 69.43 \\
AdaptFormer~\cite{chen2022adaptformer} & 0.16 & 80.56 & 84.88 & 58.83 & 72.32 \\
BitFit~\cite{zaken2022bitfit} & 0.10 & 73.30 & 78.25 & 44.10 & 62.05 \\
LoRA~\cite{hu2022lora} & 0.29 & 79.49 & 84.55 & 59.78 & 72.25 \\ 
SPT-LoRA~\cite{he2023sensitivity} & 0.43 & 81.93 & 85.93 & 61.26 & \underline{74.07} \\
E$^2$VPT~\cite{han2023e2vpt} & 0.28 & 80.01 & 84.43 & 57.39 & 71.42 \\
SA$^2$VP~\cite{pei2024sa2vp} & 0.68 & 80.97 & 85.72 & 60.80 & 73.48\\
VFPT~\cite{zeng2024visual} & 0.45 & 81.34 & 84.95 & 60.20 & 73.20\\
FPET$_{\textit{LoRA}}$~\cite{kim2025faster} & 0.30 & 80.44 & 84.60 & 61.55 & 73.36\\
FPET$_{\textit{AdaptFormer}}$~\cite{kim2025faster} & 0.17 & 80.91 & 85.83 & \textbf{61.89} & 73.75\\
VAPT~\cite{le2025adaptive} & 0.19 & 81.43 & 85.12 & 59.34 & 72.91\\
FacT~\cite{jie2023fact}	& 0.11 & 79.41	& 84.23	& 58.31	& 74.00 \\
SSF+GIST~\cite{ruan2024gist} & 0.24 & \textbf{82.56}	& \textbf{86.77}	& 60.14	& 74.01\\
\hline
\rowcolor{gray!25}GenFT & 0.27 & \underline{82.03} & \underline{86.72} & \underline{61.80} & \textbf{74.50} \\
\hline
\end{tabular}
}
\end{table}

\noindent\textbf{Pilot Validation on LLaMA-7B.} For generative task, we fine-tune LLaMA-7B~\cite{touvron2023llama} on the Alpaca dataset~\cite{taori2023alpaca}, which contains 52,000 instruction-following demonstrations generated by OpenAI's text-davinci-003. To evaluate GenFT on LLM under resource constraints, we conduct a pilot study on a randomly selected subset of 1,000 samples and compare GenFT with LoRA and PiSSA in terms of perplexity. Results are reported in Table~\ref{tab:generative}. We also observe that GenFT requires a larger learning rate (3$\times$10$^{-3}$) than LoRA and PiSSA (1$\times$10$^{-4}$) to obtain an improved learning curve, which may be attributed to its expanded rank space. This observation suggests that GenFT exhibits distinct optimization dynamics, warranting further investigation in future large-scale experiments.

\begin{table}[htbp]
\centering
\vspace{-0.8cm}
\caption{Generative task Perplexity ($\downarrow$) on LLaMA-7B.}
\label{tab:generative}
\renewcommand\arraystretch{0.8}
\setlength{\tabcolsep}{2.5mm}{
\begin{tabular}{c|ccc}
\hline
\rowcolor{gray!25}\textbf{Step} & \textbf{GenFT(2.6 M)} & \textbf{PiSSA(4.2 M)} & \textbf{LoRA(4.2 M)}\\
\hline
5 & 7.69 & \textbf{7.15} & \underline{7.64}\\
10 & \textbf{5.64} & \underline{5.84} & 7.08\\
15 & \textbf{4.45} & \underline{4.68} & 6.18\\
20 & \textbf{3.63} & \underline{3.99} & 5.27\\
25 & \textbf{3.40} & \underline{3.56} & 4.54\\
30 & \textbf{3.33} & \underline{3.38} & 3.97\\
35 & \textbf{3.23} & \underline{3.27} & 3.71\\
40 & \textbf{3.19} & \underline{3.25} & 3.62\\
\hline
\end{tabular}}
\end{table}

\textbf{Ablation Study.} GenFT comprises four main components: the layer-shared configuration ($U_{\textit{s}}, V_{\textit{s}}$), the layer-specific configuration ($A^{(\ell)}, B^{(\ell)}$), the row transformation of $W_0^{(\ell)}$, and the column transformation of $W_0^{(\ell)}$. We ablate each component individually and evaluate performance on the FGVC and VTAB-1K benchmarks. To disable the layer-shared or layer-specific module, we set the corresponding dimension to zero; for the row and column ablations, we remove the corresponding transformation. In cases where a dimension is originally zero, we set it to 8 to ensure functionality.

\begin{table}[htbp]
\centering
\vspace{-0.8cm}
\caption{Ablation on FGVC and VTAB-1K benchmarks. Details are shown in Appendix section H.}
\label{tab:ablation}
\renewcommand\arraystretch{0.8}
\setlength{\tabcolsep}{3mm}{
\begin{tabular}{c|c|cccc}
\hline
\rowcolor{gray!25}\textbf{Ablated} & \textbf{FGVC} & \multicolumn{4}{c}{\textbf{VTAB-1K}}\\
\cline{2-6}
\rowcolor{gray!25}\textbf{Model} & \textbf{Avg.} & \textbf{Nat.} & \textbf{Spe.} & \textbf{Str.} &  \textbf{Avg.}\\
\hline
w/o Share ($U_s,V_s$) & 84.22 $\downarrow\downarrow$& 77.80 & 84.40 & 39.77 & 63.18 $\downarrow\downarrow$\\
w/o Specific ($A^{(\ell)},V_s^{(\ell)}$) & 90.28 $\downarrow$& 81.10 & \underline{86.38} & 48.79 & 68.61 $\downarrow$\\
w/o Row ($F^{(\ell)}_{\textit{row}}$) & 89.96 $\downarrow$& 81.24 & 86.25 & \underline{53.36} & \underline{70.56}  $\downarrow$\\
w/o Column ($F^{(\ell)}_{\textit{col.}}$) & 89.76 $\downarrow$ & \underline{81.34} & 86.18 & 52.75 & 70.32 $\downarrow$\\ 
\hline
\rowcolor{gray!25} GenFT & \textbf{90.38} & \textbf{82.03} & \textbf{86.72} & \textbf{61.80} & \textbf{74.50}\\
\hline
\end{tabular}}
\end{table}

Results in Table~\ref{tab:ablation} show that removing any component degrades performance, with FGVC decreasing by up to 6\% and VTAB-1K by up to 11\%. In particular, removing the layer-shared configuration causes a substantial drop due to its role in cross-task generalization. Moreover, the row and column transformations, together with the rank-decomposition policy and $\sigma_1, \sigma_2$, are crucial for extracting structured information into $\Delta W^{(\ell)}$.

\begin{table}[htbp]
\centering
\caption{Time cost experiments on three costly datasets. Left is LoRA rank (r), Right is GenFT components (a+b).}
\label{tab:time}
\renewcommand\arraystretch{0.8}
\setlength{\tabcolsep}{1mm}{
\begin{tabular}{c|c|cccc}
\hline
\rowcolor{gray!25}\textbf{Method} & \textbf{Dataset} & \textbf{Time (h)} & \textbf{Param.(M)} & \textbf{Rank(r$\mid$a+b)} & \textbf{Acc.} \\
\hline
LoRA & \multirow{3}{*}{dSprites/loc} & 0.43 & 3.69 & 100 & 80.01 \\
LoRA &  & 0.41 & 0.59 & 16 & \underline{80.92} \\
GenFT &  & 0.42 & 3.69 & 100+0 & \textbf{82.07} \\
\hline
LoRA & \multirow{3}{*}{Cars} & 2.93 & 2.88 & 78 & \textbf{86.66} \\
LoRA &  & 2.81 & 0.59 & 16 & 85.31 \\
GenFT &  & 2.95 & 0.24 & 78+0 & \underline{86.16} \\
\hline
LoRA & \multirow{3}{*}{SST-2} & 2.98 & 1.92 & 52 & \underline{94.61} \\
LoRA &  & 3.01 & 0.59 & 16 & \textbf{94.84} \\
GenFT &  & 3.35 & 0.30 & 48+4 & \textbf{94.84} \\
\hline
\end{tabular}}
\end{table}

\textbf{Computational Cost Analysis.} For LoRA, we assume $\Delta W^{(\ell)} \in \mathbb{R}^{D \times D}$ with $A^{(\ell)} \in \mathbb{R}^{D \times r}$ and $B^{(\ell)} \in \mathbb{R}^{r \times D}$. The dominant per-layer cost is the matrix multiplication, i.e., $\mathcal{O}(rD^2)$; we ignore the cost of the element-wise addition, which is typically negligible compared to multiplication. For the row transformation in GenFT, we assume $W_0^{(\ell)} \in \mathbb{R}^{D \times D}$, $U_{\textit{s}} \in \mathbb{R}^{D \times a}$, and $A^{(\ell)}, B^{(\ell)} \in \mathbb{R}^{D \times b}$, where $a$ and $b$ denote the layer-shared and layer-specific dimensions, respectively. Using the simplified row transformation $F^{(\ell)}_{\textit{row}} = (W^{(\ell)}_0 U^{(\ell)}_{\textit{s}})(U_{\textit{s}}^{(\ell)})^\top + (W^{(\ell)}_0 B^{(\ell)})(A^{(\ell)})^\top,$ whose cost is dominated by matrix multiplications and is $\mathcal{O}((a+b)D^2)$ up to constant factors. The column transformation has the same order. Therefore, GenFT may introduce slightly higher training cost than LoRA when using a larger latent transformation dimension $a+b$, while both methods remain quadratic in $D$. Empirically, we select three time-consuming subsets across three benchmarks and compare LoRA and GenFT under different parameter budgets. As shown in Table~\ref{tab:time}, LoRA becomes comparable to GenFT in time cost only when using substantially more parameters, while GenFT maintains competitive or better performance with a smaller parameter budget.

We also measure peak GPU memory during training. GenFT introduces only small memory overhead compared with LoRA: 6.54~GB vs. 6.46~GB on VTAB-1K, 6.56~GB vs. 6.46~GB on FGVC, and 7.09~GB vs. 6.77~GB on GLUE. Moreover, after training, the generated update $\Delta W$ can be materialized and merged into $W_0+\Delta W$, so the additional generator cost mainly affects training rather than inference.

\textbf{Benefit from a Larger Latent Transformation Space.} We select three representative subsets from the three benchmarks for evaluation. All training hyperparameters are kept identical for LoRA and GenFT. Here, $r$ denotes the LoRA rank, while $a$ and $b$ denote the shared and layer-specific dimensions in GenFT, respectively. Results are summarized in Table~\ref{tab:benifit}. As shown, GenFT can use a larger latent transformation dimension $a+b$ with fewer trainable parameters than LoRA, leading to better or comparable performance under a small parameter budget.

\begin{table}[htbp]
\centering
\vspace{-0.8cm}
\caption{GenFT's latent transformation space V.S. LoRA's}
\label{tab:benifit}
\renewcommand\arraystretch{0.8}
\setlength{\tabcolsep}{2mm}{
\begin{tabular}{c|ccc|ccc}
\hline
\rowcolor{gray!25} & \multicolumn{3}{c|}{\textbf{LoRA}} & \multicolumn{3}{c}{\textbf{GenFT}}\\
\cline{2-7}
\rowcolor{gray!25} & \textbf{r} & \textbf{Param. (M)} & \textbf{Acc.} & \textbf{a+b} & \textbf{Param. (M)} & \textbf{Acc.}\\
\hline
\multirow{3}{*}{Cifar}& \multirow{3}{*}{84} & \multirow{3}{*}{3.10} & \multirow{3}{*}{\underline{65.79}} & 84+0& 0.26 & \textbf{71.50}\\
&  &  &  & 82+2 & 0.29 & \textbf{71.40}\\
&  &  &  & 80+4	& 0.32 & \textbf{71.25}\\
\hline
\multirow{3}{*}{CUB.} & \multirow{3}{*}{94} & \multirow{3}{*}{3.47} & \multirow{3}{*}{\underline{87.81}} & 94+0	& 0.29 & \textbf{88.97}\\
&  &  &  & 92+2 & 0.32 & \textbf{89.33}\\
&  &  &  & 90+4 & 0.35 & \textbf{88.67}\\
\hline
\multirow{3}{*}{MRPC} & \multirow{3}{*}{34} & \multirow{3}{*}{1.25} & \multirow{3}{*}{\underline{89.22}} & 34+0 & 0.11 & 88.48\\
&  &  &  & 32+2 & 0.17 & \textbf{89.95}\\
&  &  &  & 30+4 & 0.24 & 86.03\\
\hline
\end{tabular}}
\end{table}

\textbf{More Experiments Results}. To facilitate a comprehensive evaluation of GenFT, we present supplementary empirical results. Specifically, Appendix E analyzes the impact of the layer-shared dimension on final performance. Appendix F elucidates the distinctions between GenFT and LoRA regarding row and column information. Furthermore, Appendix G investigates how different initialization strategies influence model effectiveness.

\section{Conclusion}
\label{conclusion}
In this work, we propose GenFT, a $W_0$-conditioned PEFT method that uses a deterministic generator to produce task-specific updates $\Delta W$. GenFT extracts structured row- and column-wise information from $W_0$ and uses a shared-specific decomposition to balance cross-layer reuse and layer-specific flexibility. Experiments on NLP and CV benchmarks show competitive or better average performance than representative PEFT baselines, while a pilot study on LLaMA-7B suggests its feasibility for generative models.

\section{Acknowledgment}

This work is supported by RGC Research Matching Grant (RMGS-2025-01-11), Guangdong Basic and Applied Basic Research Foundation (2025A1515011674), Guangdong-Hong Kong Universities “1+1+1” Joint Research Scheme (UICR0800008-24), National Key R\&D Program of China (2024YFE0202900), and RGC GRF (12300125).

\newpage
\appendix

\section{Explanation of Finding 1}
\label{app:finding}

\textbf{Finding 1.} Given a model with $L>1$ layers, let $a$ denote the layer-shared dimension, $b$ the layer-specific dimension, and $r$ denote LoRA's low-rank dimension. Under a comparable parameter budget, GenFT can use a larger latent transformation dimension than LoRA.

\textit{Explanation.} For the $\ell$-th layer, LoRA parameterizes the update as
\begin{equation*}
\Delta W_{\text{LoRA}}^{(\ell)} = A_{\text{LoRA}}^{(\ell)} B_{\text{LoRA}}^{(\ell)},
\end{equation*}
where $A_{\text{LoRA}}^{(\ell)} \in \mathbb{R}^{D \times r}$ and $B_{\text{LoRA}}^{(\ell)} \in \mathbb{R}^{r \times D}$ are learnable parameters. Thus, the total number of LoRA parameters over $L$ layers is
\begin{equation*}
\text{Total \# of LoRA} = 2LDr .
\end{equation*}

For GenFT, the update of the $\ell$-th layer is generated from $W_0^{(\ell)}$ as
\begin{equation*}
\begin{aligned}
U^{(\ell)} &= U_s U_s^T + B^{(\ell)}(A^{(\ell)})^T,\\
F_{\textit{row}}^{(\ell)} &= \sigma_1(\textit{ratio} \cdot W_0^{(\ell)} U^{(\ell)})\odot M_p,\\
V^{(\ell)} &= V_s V_s^T + B^{(\ell)}(A^{(\ell)})^T,\\
F_{\textit{col.}}^{(\ell)} &= \sigma_2((F_{\textit{row}}^{(\ell)})^T V^{(\ell)}) \odot M_p,\\
\Delta W_{\text{GenFT}}^{(\ell)} &= \textit{scaling} \cdot F_{\textit{col.}}^{(\ell)},
\end{aligned}
\end{equation*}
where $U_s,V_s \in \mathbb{R}^{D \times a}$ are shared across layers, and $A^{(\ell)},B^{(\ell)} \in \mathbb{R}^{D \times b}$ are layer-specific parameters. Therefore, the total number of GenFT parameters is
\begin{equation*}
\text{Total \# of GenFT} = 2Da + 2LDb .
\end{equation*}

Suppose GenFT and LoRA have the same parameter count:
\begin{equation*}
2Da + 2LDb = 2LDr .
\end{equation*}
Then,
\begin{equation*}
a = L(r-b).
\end{equation*}
If $r>b$ and $L>1$, we have
\begin{equation*}
a+b-r = L(r-b)+b-r = (L-1)(r-b) > 0,
\end{equation*}
and hence
\begin{equation*}
r < a+b .
\end{equation*}
This indicates that, under the same parameter budget, GenFT can allocate a larger latent transformation dimension than LoRA. Here, $a+b$ should not be interpreted as the exact algebraic rank of $\Delta W_{\text{GenFT}}^{(\ell)}$, since nonlinear activations and masking may change the resulting matrix rank.

\section{Benchmarks Details}

Here, we introduce the details of the selected benchmarks: VTAB-1k, FGVC, and GLUE, which are widely used to evaluate the generalization and robustness of machine learning models across diverse tasks and domains.

The VTAB-1k Benchmark is designed to assess the adaptability of visual representation learning models. It comprises 19 diverse visual tasks, grouped into three categories: natural, specialized, and structured. Each task is limited to 1,000 training examples to simulate low-data scenarios, challenging models to leverage pre-trained representations effectively. Table~\ref{tab:vtab1k} provides detailed statistics and descriptions of the VTAB-1k benchmark, including task categories, and dataset sizes.

\begin{table}[htbp]
\centering
\caption{VTAB-1K Benchmark~\cite{zhai2019large} Details}
\label{tab:vtab1k}
\renewcommand\arraystretch{1.2} 
\resizebox{\linewidth}{!}{
\begin{tabular}{clcccc}
\hline
\rowcolor{gray!25}Category & Dataset & \# Classes & Train & Val & Test \\
\hline
\multirow{7}{*}{\textit{Natural}} & CIFAR100~\cite{krizhevsky2009learning} & 100 & \multirow{7}{*}{800/1,000} & \multirow{7}{*}{200} & 10,000 \\
 & Caltech101~\cite{fei2004learning} & 102 & & & 6,084 \\
 & DTD~\cite{cimpoi2014describing} & 47 &  &  & 1,880 \\
 & Oxford-Flowers102~\cite{nilsback2006visual} & 102 & & & 6,149 \\
 & Oxford-Pets~\cite{parkhi2012cats} & 37 & &  & 3,669 \\
 & SVHN~\cite{netzer2011reading} & 10 & & & 26,032 \\
 & Sun397~\cite{xiao2010sun} & 397 & & & 21,750 \\
\hline
\multirow{4}{*}{\textit{Specialized}} & Patch Camelyon~\cite{veeling2018rotation} & 2 & \multirow{4}{*}{800/1,000} & \multirow{4}{*}{200} & 32,768\\
 & EuroSAT~\cite{helber2019eurosat} & 10 &  &  & 5,400 \\
 & Resisc45~\cite{cheng2017remote} & 45 &  &  & 6,300 \\
 & Retinopathy~\cite{diabetic-retinopathy-detection} & 5 & & & 42,670 \\
\hline
\multirow{8}{*}{\textit{Structured}} & Clevr-count~\cite{johnson2017clevr} & 8 & \multirow{8}{*}{800/1,000} & \multirow{8}{*}{200} & 15,000\\
 & Clevr-distance~\cite{johnson2017clevr} & 6 & & & 15,000 \\
 & DMLab~\cite{beattie2016deepmind} & 6 &  &  & 22,735 \\
 & KITTI-Dist~\cite{geiger2013vision} & 4 &  &  & 711 \\
 & dSprites/location~\cite{higgins2017beta} & 16 & & & 73,728 \\
 & dSprites/orientation~\cite{higgins2017beta} & 16 & & & 73,728 \\
 & SmallNORB/azimuth~\cite{lecun2004learning} & 18 & & & 12,150 \\
 & SmallNORB/elevation~\cite{lecun2004learning} & 18 & & & 12,150 \\
\hline
\end{tabular}
}
\end{table}

The FGVC benchmark focuses on tasks requiring fine-grained discrimination within specific categories, such as identifying species of birds, models of cars, or types of aircraft. FGVC datasets typically involve high intra-class similarity and inter-class variability, making them challenging for visual recognition systems. Notable datasets include CUB-200-2011 (birds), NABirds, Flowers102, Stanford Cars, and Stanford Dogs. Table~\ref{tab:fgvc_details} summarizes the FGVC benchmark, including dataset sizes and number of classes.

\begin{table}[htbp]
\centering
\caption{FGVC Benchmark~\cite{jia2022visual} Details}
\label{tab:fgvc_details}
\renewcommand\arraystretch{1.2} 
\setlength{\tabcolsep}{6mm} 
{
\begin{tabular}{lccccc}
\hline
\rowcolor{gray!25}Dataset & \# Classes & Train & Val & Test \\
\hline
CUB-200-2011~\cite{wah2011caltech} & 200 & 5,394 & 600 & 5,794\\
NABirds~\cite{van2015building} & 555 & 21,536 & 2,393 & 24,633\\
Flowers102~\cite{nilsback2008automated} & 102 & 1,020 & 1,020 & 6,149\\
Stanford Dogs~\cite{khosla2011novel} & 120 & 10,800 & 1,200 & 8,580\\
Stanford Cars~\cite{krause20133d} & 196 & 7,329 & 815 & 8,041\\
\hline
\end{tabular}
}
\end{table}

The GLUE benchmark is a collection of nine natural language understanding (NLU) tasks aimed at evaluating models' ability to generalize across diverse linguistic tasks. These tasks cover single-sentence classification, similarity and paraphrase detection, and natural language inference (NLI), with varying dataset sizes and domains to test sample-efficient learning and cross-task knowledge transfer. Table~\ref{tab:glue} presents the details of the GLUE benchmark, including training and test set sizes, task, task types, evaluation metrics.

\begin{table}[htbp]
\centering
\caption{GLUE Benchmark~\cite{wangglue} Details}
\label{tab:glue}
\renewcommand\arraystretch{1.2} 
\resizebox{\linewidth}{!}{
\begin{tabular}{lccccc}
\hline
\rowcolor{gray!25}Corpus & $|$Train$|$ & $|$Test$|$ & Task & Task Type & Adopted Metrics \\
\hline
\multicolumn{6}{c}{\textit{Single-Sentence Tasks}} \\
\hline
CoLA & 8.5k & 1k & acceptability & Classification & Matthews Corr.\\
SST-2 & 67k & 1.8k & sentiment & Classification & ACC. \\
\hline
\multicolumn{6}{c}{\textit{Similarity and Paraphrase Tasks}} \\
\hline
MRPC & 3.7k & 1.7k & paraphrase & Classification & ACC.\\
STS-B & 7k & 1.4k & sentence similarity & Regression  & Pearson\\
QQP & 364k & 391k & paraphrase & Classification  & ACC. \\
\hline
\multicolumn{6}{c}{\textit{Inference Tasks}} \\
\hline
MNLI & 393k & 20k & NLI & Classification  & ACC.\\
QNLI & 105k & 5.4k & QA/NLI & Classification  & ACC. \\
RTE & 2.5k & 3k & NLI & Classification  & ACC. \\
WNLI & 634 & 146 & coreference/NLI & Classification  & ACC. \\
\hline
\end{tabular}
}
\end{table}

\section{Experiments Hyperparameter Details}

The experiments across the VTAB-1K, FGVC, and GLUE benchmarks utilize a comprehensive set of hyperparameters to fine-tune model performance and training efficiency. Under the \textbf{GenFT} category, hyperparameters such as $W_0$ Ratio, Initial method of $A$ (Init. $A$), Initial method of $B$ (Init. $B$), $\sigma_1$ and $\sigma_2$ Activations, Shared and Specific Dimensions, Bias, Dropout, and Scaling control the feature transformation and adaptation processes, ensuring the model effectively captures task-specific patterns while maintaining generalization. The \textbf{Training} category includes hyperparameters like Seed, Optimizer, Number of GPUs, GPU Memory, Label Smoothing, Batch Size, Learning Rate, Weight Decay, Epochs, Warmup Epochs, Cycle Decay, and Training Time, which govern the optimization process, resource allocation, and training dynamics to achieve stable and efficient convergence. Detailed hyperparameter configurations for each benchmark are provided in their respective tables: Table~\ref{tab:hyper_vtab1k} for VTAB-1K, Table~\ref{tab:hyper_fgvc} for FGVC, and Table~\ref{tab:hyper_glue} for GLUE. 

For initialization methods, we search Kaiming Uniform (K-U), Xavier Uniform (X-U), Normal (N), and Zeros (Z). For activation functions, we try ReLU (R), LeakyReLU (LR), Tanh (T), GeLU (G), and Identity (I).

\begin{table}[hbtp]
\centering
\caption{Hyperparameters and computing resources on VTAB-1K benchmark.}
\label{tab:hyper_vtab1k}
\renewcommand\arraystretch{1.0} 
\resizebox{\linewidth}{!}{
\begin{tabular}{llccccccccccccccccccc}
\hline
\rowcolor{gray!25}& \textbf{Hyperparam} & \rotatebox{90}{\textbf{CIFAR-100}} & \rotatebox{90}{\textbf{Caltech101}} & \rotatebox{90}{\textbf{DTD}} & \rotatebox{90}{\textbf{Flowers102}} & \rotatebox{90}{\textbf{Pets}} & \rotatebox{90}{\textbf{SVHN}} & \rotatebox{90}{\textbf{Sun397}} & \rotatebox{90}{\textbf{Patch Camelyon}} & \rotatebox{90}{\textbf{EuroSAT}} & \rotatebox{90}{\textbf{Resisc45}} & \rotatebox{90}{\textbf{Retinopathy}} & \rotatebox{90}{\textbf{Clevr/count}} & \rotatebox{90}{\textbf{Clevr/distance}} & \rotatebox{90}{\textbf{DMLab}} & \rotatebox{90}{\textbf{KITTI/distance}} & \rotatebox{90}{\textbf{dSprites/loc}} & \rotatebox{90}{\textbf{dSprites/ori}} & \rotatebox{90}{\textbf{SmallNORB/azi}} & \rotatebox{90}{\textbf{SmallNORB/ele}} \\
\hline
\multirow{10}{*}{\rotatebox{90}{\textbf{GenFT}}}& $W_0$ Ratio & 1.0 & 1.0 & 0.6 & 0.4 & 1.0 & 1.0 & 0.6 & 1.0 & 1.2 & 1.0 & 1.0 & 1.0 & 1.4 & 1.0 & 1.4 & 1.0 & 1.0 & 1.0 & 1.0 \\
& Init. $A$ & K-U & K-U & N & X-U & Z & Z & N & X-U & K-U & K-U & X-U & Z & Z & N & Z & Z & Z & Z & K-U \\
& Init. $B$ & K-U & K-U & X-U & Z & Z & Z & X-U & K-U & X-U & K-U & X-U & N & Z & N & N & Z & Z & Z & N \\
& $\sigma_1$ Activation & LR & I & I & LR & G & I & I & T & I & I & I & I & I & I & I & I & I & I & I \\
& $\sigma_2$ Activation & I & I & R & T & LR & I & I & T & LR & I & LR & T & LR & I & LR & I & I & I & T \\
& Shared Dim. & 82 & 82 & 28 & 48 & 31 & 100 & 44 & 100 & 100 & 100 & 100 & 88 & 32 & 100 & 82 & 100 & 80 & 100 & 32 \\
& Specific Dim. & 2 & 2 & 2 & 6 & 2 & 0 & 6 & 0 & 0 & 0 & 1 & 1 & 5 & 2 & 1 & 0 & 0 & 0 & 5 \\
& Bias & T & T & T & T & T & T & T & T & T & T & T & T & F & T & T & T & F & T & T \\
& Dropout & 0.1 & 0.1 & 0.1 & 0.0 & 0.1 & 0.1 & 0.05 & 0.1 & 0.1 & 0.1 & 0.1 & 0.1 & 0.1 & 0.1 & 0.1 & 0.1 & 0.1 & 0.1 & 0.1 \\
& Scaling & 1.0 & 1.0 & 0.5 & 0.1 & 0.1 & 1.0 & 0.7 & 0.9 & 1.2 & 1.0 & 1.2 & 1.2 & 1.0 & 1.2 & 1.0 & 1.0 & 1.0 & 0.8 & 1.0 \\
\hline
\multirow{12}{*}{\rotatebox{90}{\textbf{Training}}}& Seed & \multicolumn{19}{c}{42} \\
& Optimizer & \multicolumn{19}{c}{AdamW} \\
& Cycle Decay & \multicolumn{19}{c}{0.1} \\
& Batch Size & \multicolumn{19}{c}{64} \\
& Warmup Epochs & \multicolumn{19}{c}{10} \\
& \# GPUs & \multicolumn{19}{c}{1} \\
& GPU Memory (G) & 6.5 & 6.5 & 6.6 & 6.5 & 6.6 & 6.5 & 6.5 & 6.6 & 6.6 & 6.5 & 6.6 & 6.6 & 6.6 & 6.5 & 6.6 & 6.5 & 6.5 & 6.5 & 6.6 \\
& Training Time (Min) & 10 & 9 & 9 & 10 & 9 & 17 & 13 & 16 & 9 & 9 & 20 & 12 & 11 & 14 & 10 & 25 & 25 & 10 & 10 \\
& Label Smooth & 0.1 & 0.1 & 0.2 & 0.0 & 0.2 & 0.1 & 0.2 & 0.1 & 0.1 & 0.1 & 0.1 & 0.1 & 0.0 & 0.1 & 0.1 & 0.0 & 0.0 & 0.0 & 0.0 \\
& Learning Rate ($10^{-3}$) & 5.0 & 1.0 & 1.0 & 1.0 & 1.0 & 1.0 & 1.0 & 1.0 & 1.0 & 1.0 & 1.0 & 1.0 & 1.0 & 1.0 & 1.0 & 1.0 & 1.0 & 1.0 & 1.0 \\
& Weight Decay ($10^{-4}$) & 1.0 & 1.0 & 1.0 & 1.0 & 0.0 & 1.0 & 1.0 & 1.0 & 1.0 & 1.0 & 1.0 & 1.0 & 1.0 & 1.0 & 1.0 & 1.0 & 1.0 & 1.0 & 0.0 \\
& Epoch & 100 & 100 & 100 & 100 & 100 & 120 & 100 & 100 & 100 & 100 & 100 & 100 & 100 & 100 & 100 & 100 & 100 & 100 & 100 \\
\hline
\end{tabular}
}
\end{table}

\begin{table}[hbtp]
\centering
\caption{Hyperparameters and computing resources on the FGVC benchmark.}
\label{tab:hyper_fgvc}
\renewcommand\arraystretch{1.2} 
\resizebox{\linewidth}{!}{
\begin{tabular}{llccccc}
\hline
\rowcolor{gray!25}& \textbf{Hyperparam} & \textbf{CUB-200-2011} & \textbf{NABirds} & \textbf{Oxford Flowers} & \textbf{Stanford Dogs} & \textbf{Stanford Cars} \\
\hline
\multirow{10}{*}{\rotatebox{90}{\textbf{GenFT}}}& $W_0$ Ratio & 1.6 & 1.0 & 0.8 & 0.6 & 1.4 \\
& Init. $A$ & N & X-U & X-U & K-U & N \\
& Init. $B$ & Z & X-U & X-U & Z & Z \\
& $\sigma_1$ Activation & T & I & G & LR & LR \\
& $\sigma_2$ Activation & LR & I & G & LR & I \\
& Shared Dim. & 92 & 82 & 100 & 82 & 78 \\
& Specific Dim. & 2 & 2 & 2 & 1 & 0 \\
& Bias & T & T & T & T & T \\
& Dropout & 0.2 & 0.2 & 0.1 & 0.0 & 0.1 \\
& Scaling & 0.4 & 1.0 & 1.0 & 0.4 & 1.6 \\
\hline
\multirow{12}{*}{\rotatebox{90}{\textbf{Training}}}& Seed & \multicolumn{5}{c}{42} \\
& Optimizer & \multicolumn{5}{c}{AdamW} \\
& Cycle Decay & \multicolumn{5}{c}{0.1} \\
& Batch Size & \multicolumn{5}{c}{64} \\
& Warmup Epochs & \multicolumn{5}{c}{10} \\
& \# GPUs & \multicolumn{5}{c}{1} \\
& GPU Memory (G) & 6.6 & 6.5 & 6.6 & 6.6 & 6.5 \\
& Training Time (Min) & 93 & 769 & 28 & 152 & 176 \\
& Label Smooth & 0.1 & 0.1 & 0.1 & 0.1 & 0.1 \\
& Learning Rate ($10^{-3}$) & 0.8 & 1.0 & 3.0 & 1.0 & 0.8 \\
& Weight Decay ($10^{-4}$) & 1.0 & 1.0 & 1.0 & 1.0 & 1.0 \\
& Epoch & 100 & 100 & 100 & 100 & 100 \\
\hline
\end{tabular}
}
\end{table}

\begin{table}[hbtp]
\centering
\caption{Hyperparameters and computing resources on the GLUE benchmark.}
\label{tab:hyper_glue}
\scriptsize
\renewcommand\arraystretch{1.2}
\setlength{\tabcolsep}{1mm}
{
\begin{tabular}{llcccccccc}
\hline
\rowcolor{gray!25}& \textbf{Hyperparameter} & \textbf{CoLA} & \textbf{SST-2} & \textbf{MRPC} & \textbf{STS-B} & \textbf{QQP} & \textbf{MNLI} & \textbf{QNLI} & \textbf{RTE} \\
\hline
\multirow{10}{*}{\rotatebox{90}{\textbf{GenFT}}}& $W_0$ Ratio & 0.05 & 0.3 & 0.1 & 0.01 & 0.2 & 0.2 & 0.5 & 1.6 \\
& Init. $A$ & K-U & X-U & Z & Z & K-U & K-U & K-U & Z \\
& Init. $B$ & K-U & X-U & Z & Z & K-U & K-U & K-U & Z \\
& $\sigma_1$ Activation & I & T & LR & R & LR & LR & I & LR \\
& $\sigma_2$ Activation & I & LR & LR & I & I & I & I & G \\
& Shared Dim. & 72 & 48 & 32 & 48 & 48 & 32 & 24 & 76 \\
& Specific Dim. & 2 & 4 & 2 & 6 & 0 & 2 & 2 & 2 \\
& Bias & F & F & F & F & F & F & F & F \\
& Dropout & 0.05 & 0.1 & 0.0 & 0.0 & 0.05 & 0.05 & 0.1 & 0.1 \\
& Scaling & 0.05 & 0.05 & 0.1 & 0.001 & 0.2 & 0.2 & 0.05 & 0.15 \\
\hline
\multirow{10}{*}{\rotatebox{90}{\textbf{Training}}}& Seed & \multicolumn{8}{c}{42} \\
& Optimizer & \multicolumn{8}{c}{AdamW} \\
& Warmup Ratio & \multicolumn{8}{c}{0.06} \\
& Batch Size & 64 & 32 & 64 & 64 & 64 & 64 & 64 & 64 \\
& \# GPUs & \multicolumn{8}{c}{1} \\
& GPU Memory (G) & 2.0 & 1.7 & 3.6 & 4.1 & 9.4 & 12.6 & 14.4 & 8.9 \\
& Training Time (Min) & 13 & 201 & 8 & 12 & 709 & 1013 & 295 & 13 \\
& Learning Rate ($10^{-4}$) & 10.0 & 3.0 & 3.0 & 3.0 & 3.0 & 3.0 & 5.0 & 4.0 \\
& Weight Decay ($10^{-4}$) & 10.0 & 1.0 & 1.0 & 1.0 & 0.0 & 0.0 & 5.0 & 20.0 \\
& Epoch & 100 & 100 & 100 & 100 & 100 & 100 & 100 & 100 \\
\hline
\end{tabular}
}
\end{table}

\section{Results on FGVC}

\textbf{Implementation Details.} We adopt the ViT-Base model~\cite{dosovitskiy2020image} (85.8M parameters) for multi-class and fine-grained image classification. We evaluate GenFT on VTAB-1K~\cite{zhai2019large}, which contains 19 sub-datasets spanning natural, specialized, and structured images, and on FGVC, including CUB200-2011 (CUB200)~\cite{wah2011caltech}, NABirds~\cite{van2015building}, Flowers102 (Flowers)~\cite{nilsback2008automated}, Stanford Dogs (Dogs)~\cite{khosla2011novel}, and Stanford Cars (Cars)~\cite{krause20133d}. We report top-1 accuracy as the primary evaluation metric. Our experimental setup follows V-PETL Bench~\cite{xin2024v}. We conduct random hyperparameter search over the layer-shared dimension $a$, layer-specific dimension $b$, activation functions $\sigma_1$ and $\sigma_2$, and initialization schemes for $A_{\textit{specific}}$ and $B_{\textit{specific}}$, using the AdamW optimizer with a preset weight decay rate. To ensure reproducibility, we fix all random seeds.

\begin{table}[htbp]
\centering
\caption{Experimental results on FGVC benchmark~\cite{jia2022visual}. Its backbone is same as VTAB-1K.}
\label{tab:fgvc}
\renewcommand\arraystretch{1.0}
\resizebox{\linewidth}{!}{
\begin{tabular}{c|cccccccc}
\hline
\rowcolor{gray!25}\textbf{Method} & \makecell{\textbf{Param.}\\\textbf{(M)}} & \makecell{\textbf{CUB.} \\\textbf{(Acc.)}} & \makecell{\textbf{NAB.} \\\textbf{(Acc.)}} & \makecell{\textbf{Flo.} \\\textbf{(Acc.)}} & \makecell{\textbf{Dogs} \\\textbf{(Acc.)}} & \makecell{\textbf{Cars} \\\textbf{(Acc.)}} & \textbf{Avg.}\\
\hline
Full FT~\cite{jia2022visual} & 85.88 & 87.3 & 82.7 & 98.8 & 89.4 & 84.5 & 88.54 \\
Linear~\cite{jia2022visual} & 0.00 & 85.3 & 75.9 & 97.9 & 86.2 & 51.3 & 79.32 \\
Adapter~\cite{houlsby2019parameter} & 0.16 & 87.1 & 84.3 & 98.5 & 89.8 & 68.6 & 85.66 \\
VPT~\cite{jia2022visual} & 0.56 & 88.5 & 84.2 & 99.0 & 90.2 & 83.6 & 89.10 \\
AdaptFormer~\cite{chen2022adaptformer} & 0.16 & 88.4 & 84.7 & 99.2 & 88.2 & 81.9 & 88.48 \\
BitFit~\cite{zaken2022bitfit} & 0.10 & 87.7 & 85.2 & 99.2 & 86.5 & 81.5 & 88.02 \\
LoRA~\cite{hu2022lora} & 0.43 & 85.6 & 79.8 & 98.9 & 87.6 & 72.0 & 84.78 \\
SPT-LoRA~\cite{he2023sensitivity} & 0.43 & 88.6 & 83.4 & \textbf{99.5} & 91.4 & \textbf{87.3} & 90.04\\
E$^2$VPT~\cite{han2023e2vpt} & 0.28 & 89.1 & 84.6 & 99.1 & 90.5 & 82.8 & 89.22\\
SA$^2$VP~\cite{pei2024sa2vp} & 0.68 & 89.1 & \textbf{85.8} & \underline{99.3} & \textbf{92.1} & 84.1 & \underline{90.08} \\
VFPT~\cite{zeng2024visual}  & 0.45 & 88.7 & 84.5 & 99.1 & 90.4 & 83.6 & 89.24 \\
VAPT~\cite{le2025adaptive} & 0.19 & \textbf{89.7} & 84.6 & 99.1 & \underline{91.7} & 82.8 & 89.58 \\
\hline
\rowcolor{gray!25}GenFT & 0.27 & \underline{89.3} & \underline{85.6} & \underline{99.3} & 91.5 & \underline{86.2} & \textbf{90.38} \\
\hline
\end{tabular}
}
\end{table}

\textbf{Results on FGVC.} Results are presented in Table~\ref{tab:fgvc}. GenFT ranks second on CUB200, NABirds, Flowers, and Cars, and third on Dogs, while achieving the highest average accuracy (90.38\%) across the benchmark. With only 0.29M parameters, GenFT leverages layer-shared and layer-specific information via the rank-decomposition policy and extracts structured information from $W_0$ through row and column transformations with activation functions. This design yields robust generalization and stable performance across fine-grained classification tasks.

\section{Layer-shared Dimension Analysis.} 
\begin{figure}[ht!]
\centering
\begin{subfigure}[b]{0.45\linewidth}
    \centering
    \includegraphics[scale=0.18]{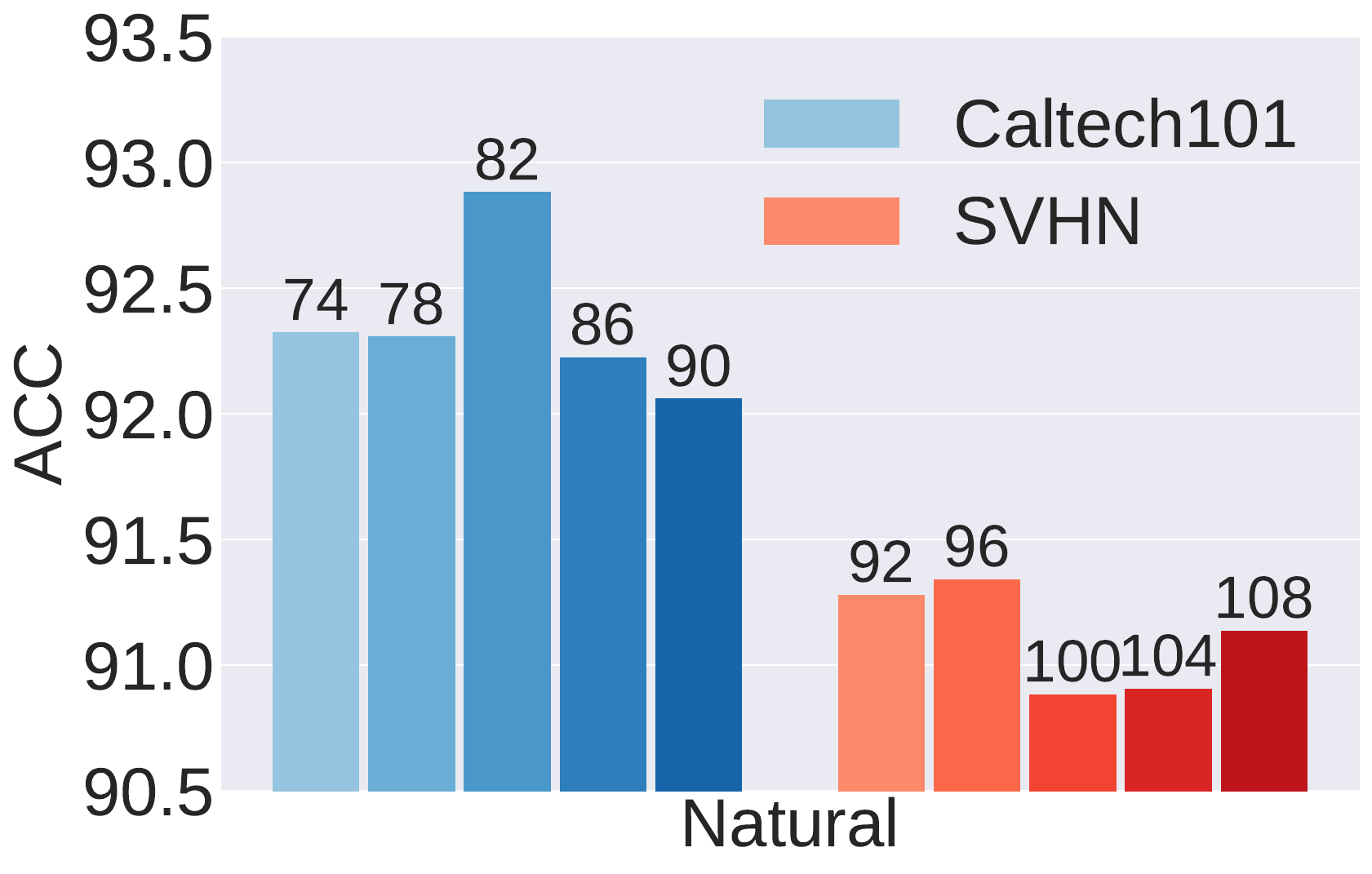}
    \caption{Natural image on VTAB-1K}
\end{subfigure}\hfill
\begin{subfigure}[b]{0.45\linewidth}
    \centering
    \includegraphics[scale=0.18]{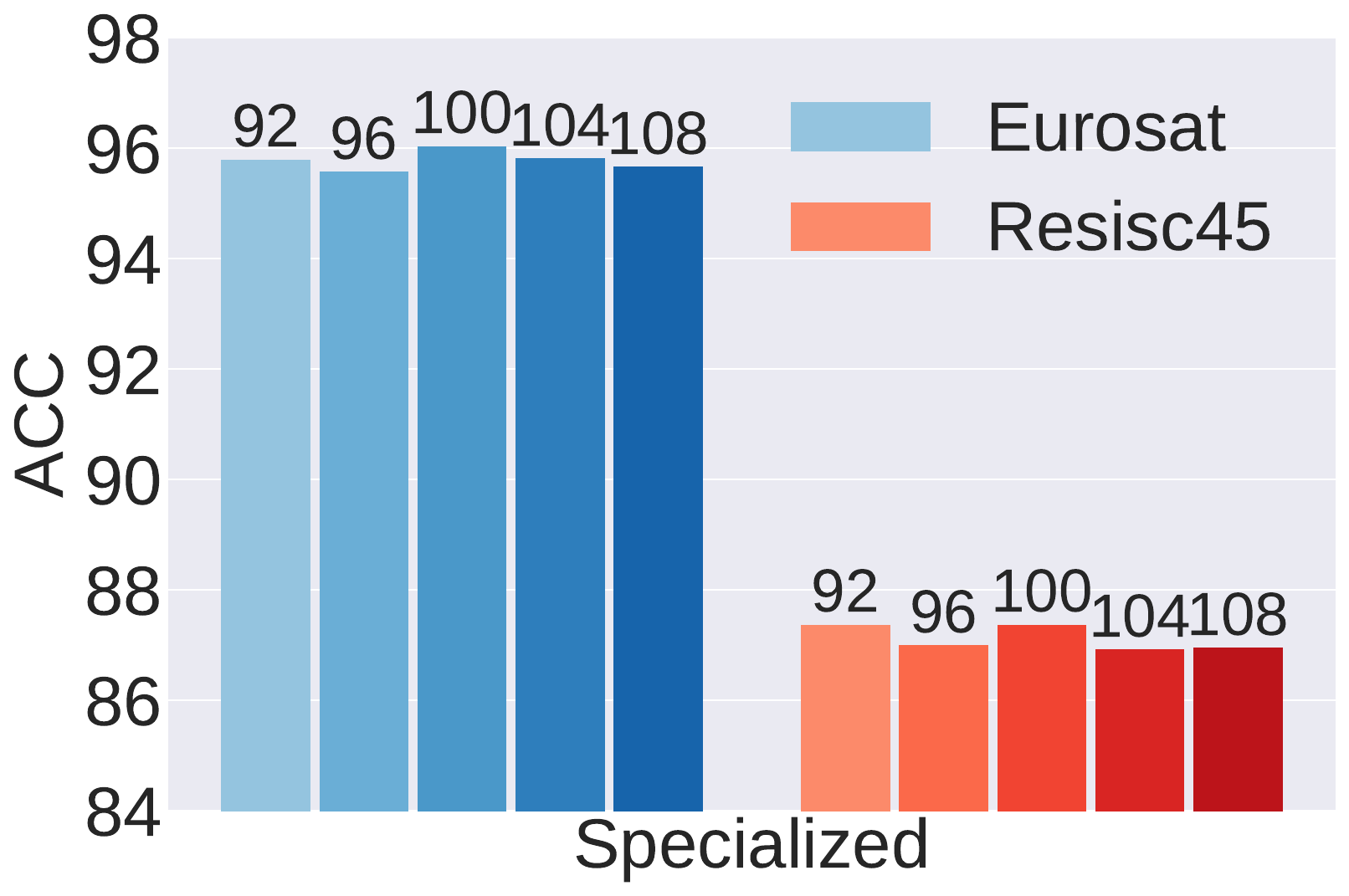}
    \caption{Specialized image on VTAB-1K}
\end{subfigure}\hfill
\begin{subfigure}[b]{0.45\linewidth}
    \centering
    \includegraphics[scale=0.18]{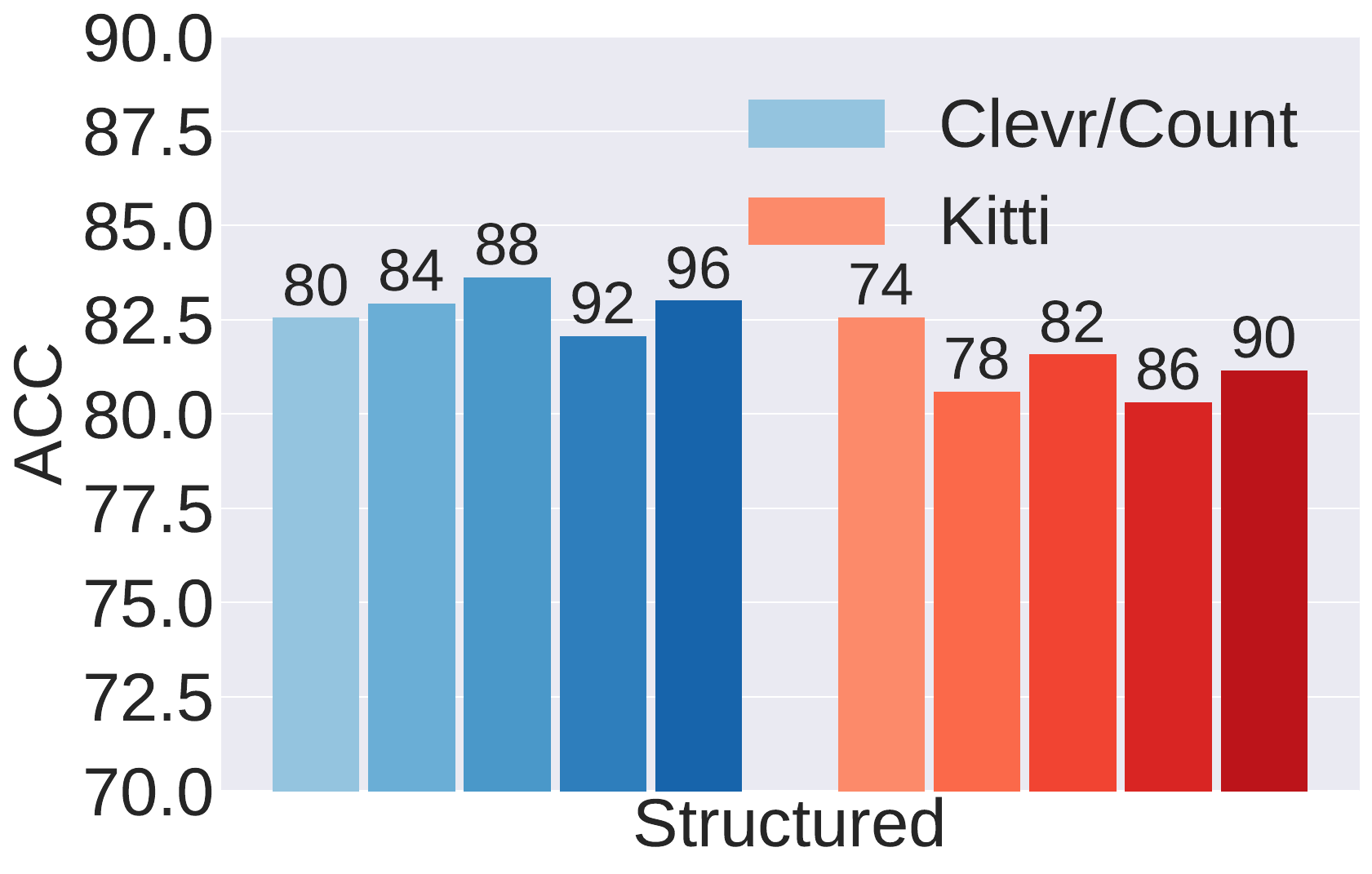}
    \caption{Structured image on VTAB-1K}
\end{subfigure}
\caption{The performance of layer-shared dimension variation.}
\label{fig:layer-shared}
\end{figure}

Motivated by the ablation results showing that removing the layer-shared dimension substantially degrades performance, we evaluate GenFT with varying layer-shared dimensions on VTAB-1K and report accuracy. Fig.~\ref{fig:layer-shared} shows the performance trends, where bar labels indicate the corresponding dimensions. For specialized images, a shared dimension of 82 is critical for Caltech101, improving accuracy by about 1\%, whereas SVHN benefits from a smaller dimension. Overall, performance is less sensitive to the shared dimension on specialized images. In contrast, for structured images, Clevr/Count and Kitti exhibit larger performance variations as the shared dimension changes, suggesting higher sensitivity due to their reliance on structured patterns. GenFT’s rank-decomposition policy helps adapt the layer-shared dimension to such structured image tasks.

\vspace{-1cm}
\section{Row and Column Information} 

\begin{figure}[ht!]
    \centering
    \begin{subfigure}{0.44\textwidth}
        \centering
        \includegraphics[scale=0.48]{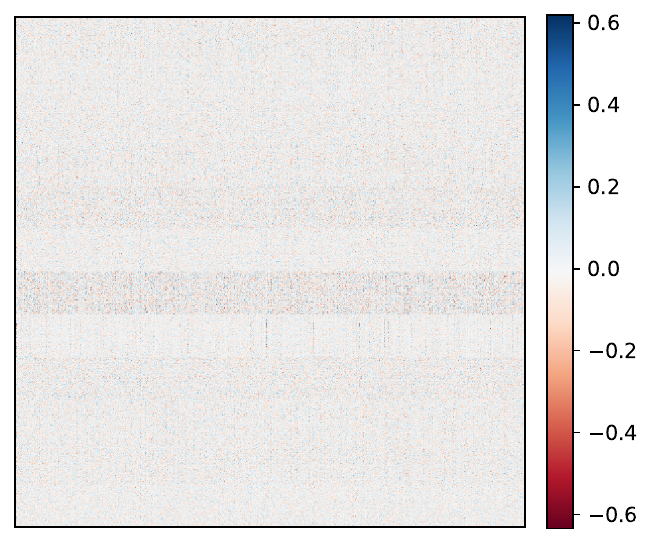}
        \caption{Pretrained $W_Q$}
    \end{subfigure}
    \hfill 
    \begin{subfigure}{0.44\textwidth}
        \centering
        \includegraphics[scale=0.48]{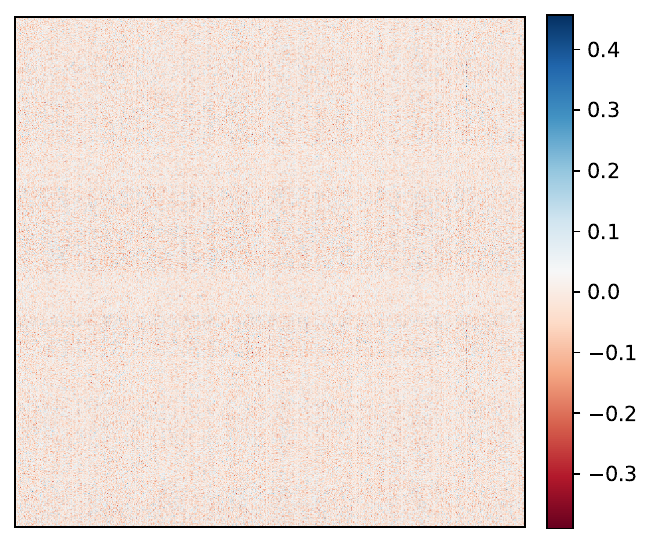}
        \caption{Pretrained $W_V$}
    \end{subfigure}
    \caption{ViT pretrained $W_0$ at layer 2}
    \label{fig:pretrained}
\end{figure}

\begin{figure}[ht!]
    \centering
    \begin{subfigure}{0.44\textwidth}
        \centering
        \includegraphics[scale=0.48]{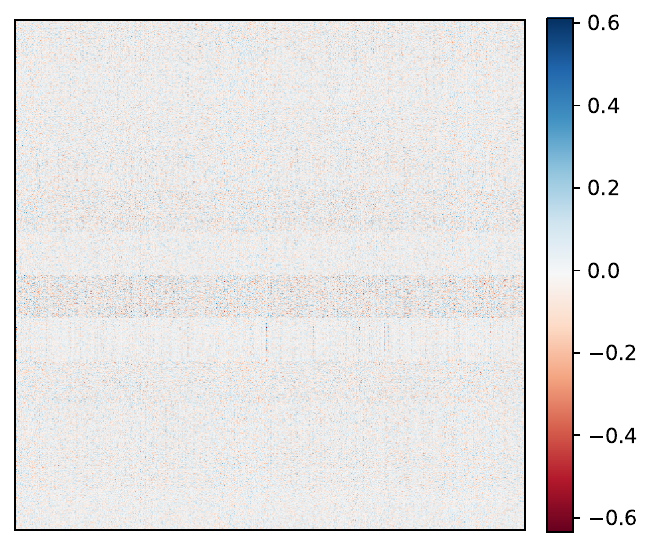}
        \caption{LoRA $W_Q + \Delta W_Q$}
    \end{subfigure}
    \hfill 
    \begin{subfigure}{0.44\textwidth}
        \centering
        \includegraphics[scale=0.48]{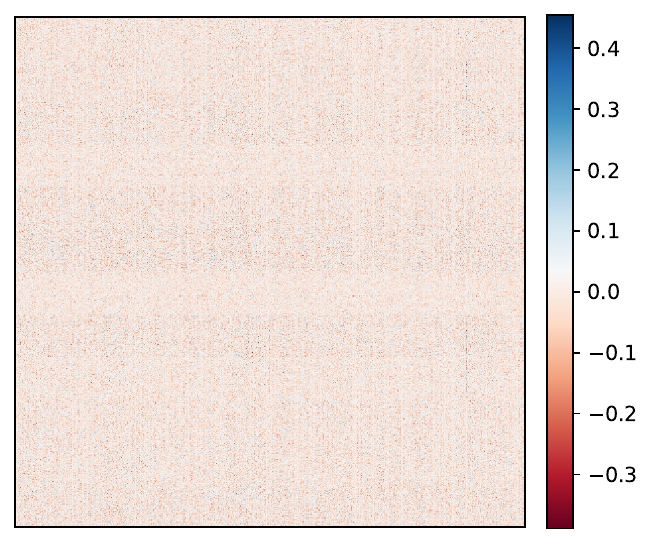}
        \caption{LoRA $W_V + \Delta W_V$}
    \end{subfigure}
    \caption{ViT pretrained $W_0$ at layer 2}
    \label{fig:lora}
\end{figure}

\begin{figure}[ht!]
    \centering
    \begin{subfigure}{0.44\textwidth}
        \centering
        \includegraphics[scale=0.48]{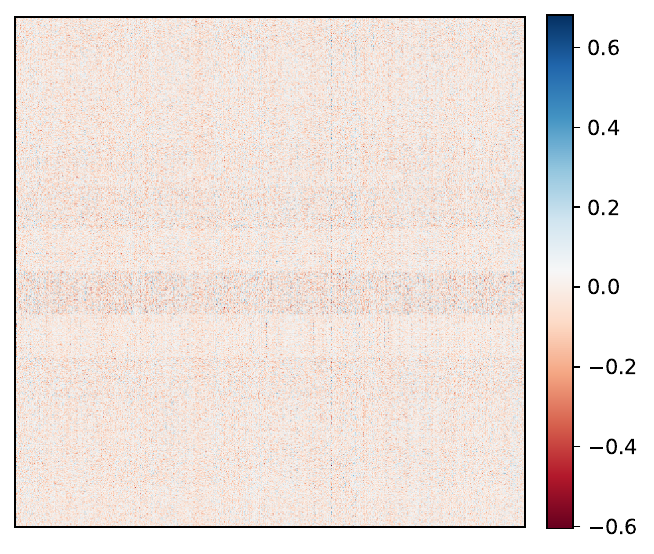}
        \caption{GenFT $W_Q + \Delta W_Q$}
    \end{subfigure}
    \hfill 
    \begin{subfigure}{0.44\textwidth}
        \centering
        \includegraphics[scale=0.48]{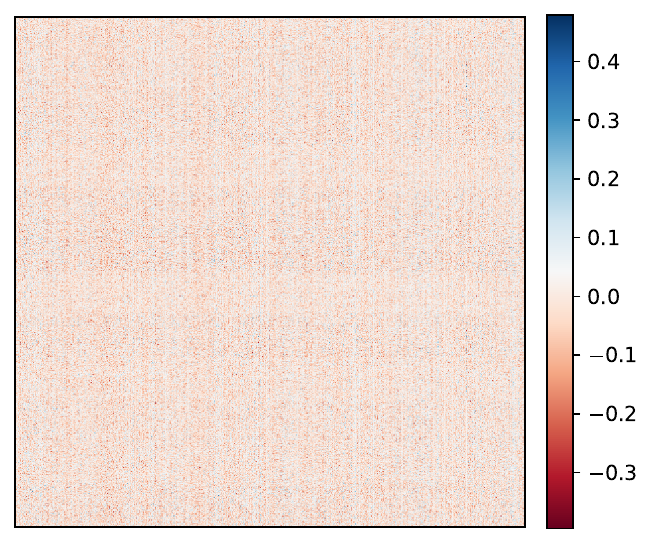}
        \caption{GenFT $W_V + \Delta W_V$}
    \end{subfigure}
    \caption{ViT pretrained $W_0$ at layer 2}
    \label{fig:genft}
\end{figure}

\begin{figure}[ht!]
    \centering
    \begin{subfigure}{0.44\textwidth}
        \centering
        \includegraphics[scale=0.18]{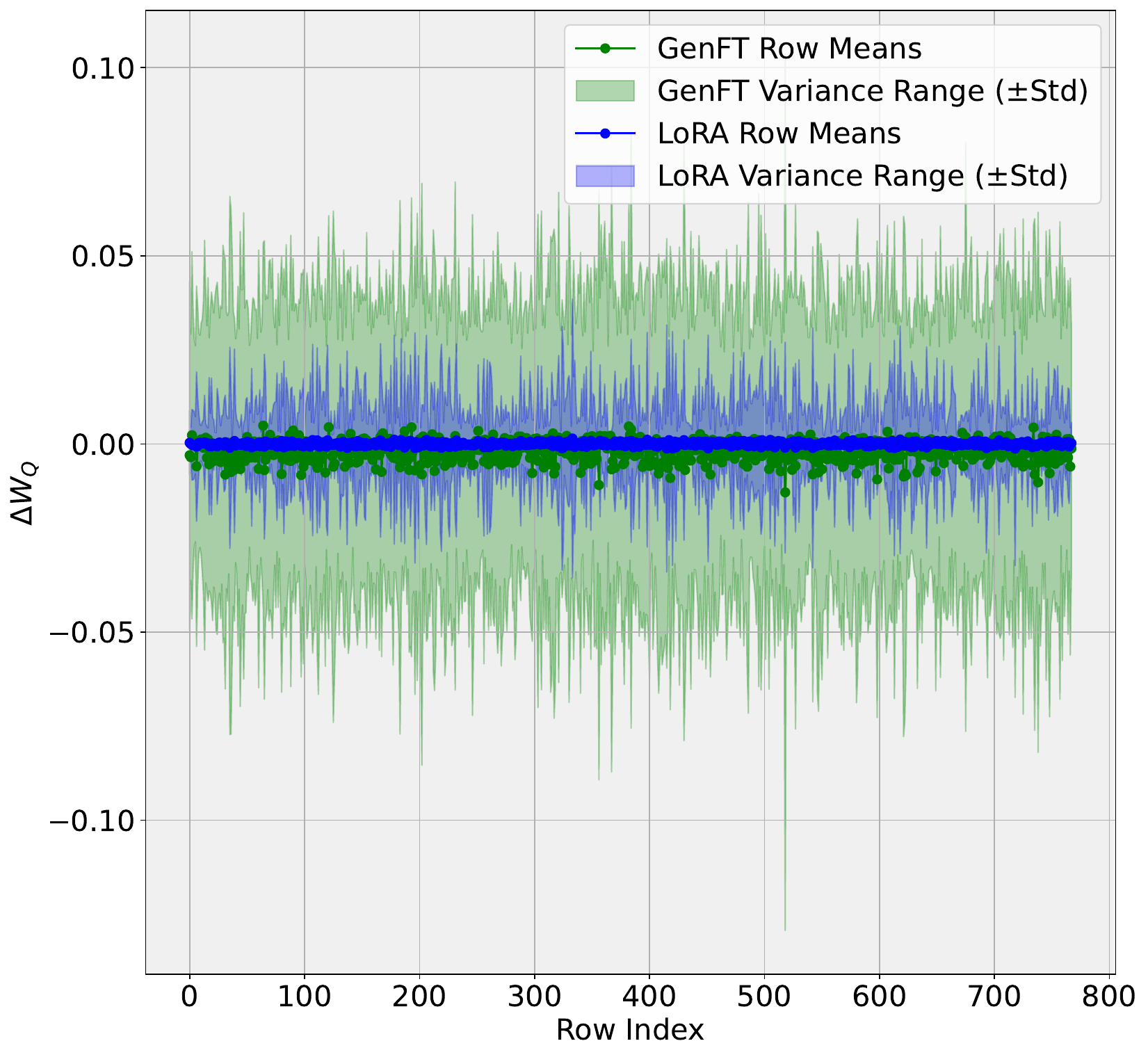}
        \caption{GenFT $\Delta W_Q$ vs LoRA $\Delta W_Q$}
    \end{subfigure}
    \hfill 
    \begin{subfigure}{0.44\textwidth}
        \centering
        \includegraphics[scale=0.18]{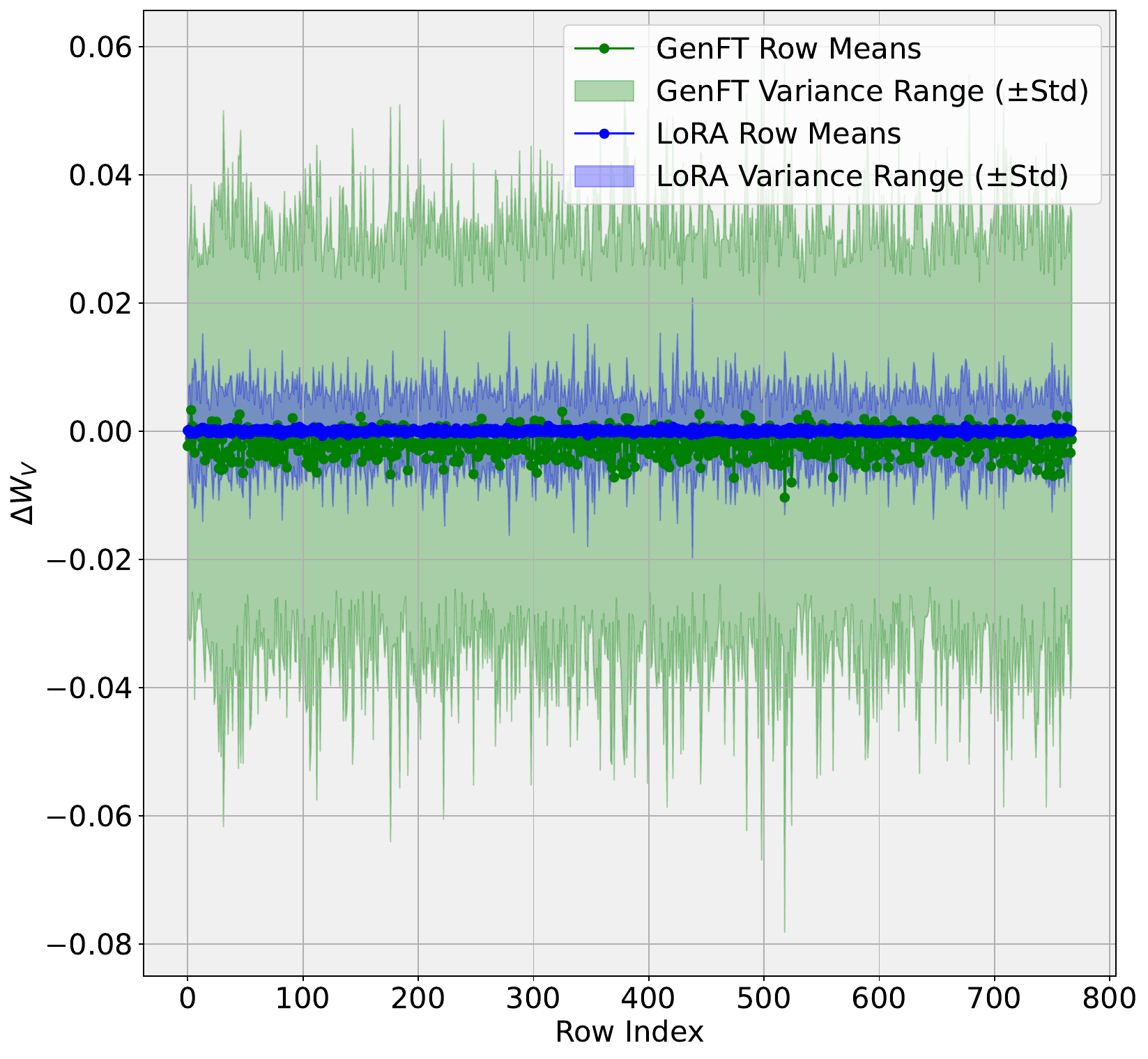}
        \caption{GenFT $\Delta W_V$ vs LoRA $\Delta W_V$}
    \end{subfigure}
    \caption{Row information comparison}
    \label{fig:genft_lora_row}
\end{figure}

\begin{figure}[ht!]
    \centering
    \begin{subfigure}{0.44\textwidth}
        \centering
        \includegraphics[scale=0.18]{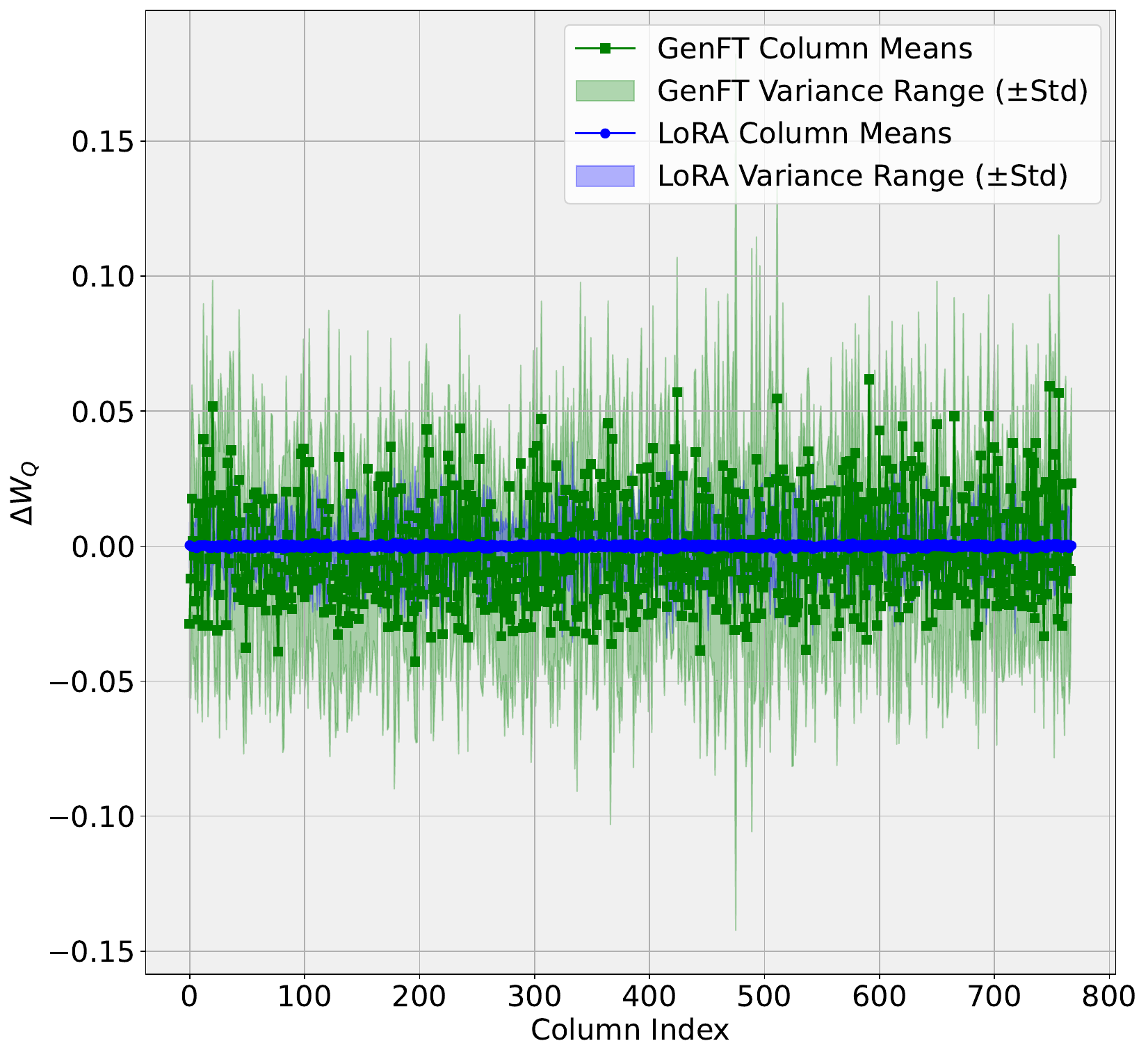}
        \caption{GenFT $\Delta W_Q$ vs LoRA $\Delta W_Q$}
    \end{subfigure}
    \hfill 
    \begin{subfigure}{0.44\textwidth}
        \centering
        \includegraphics[scale=0.18]{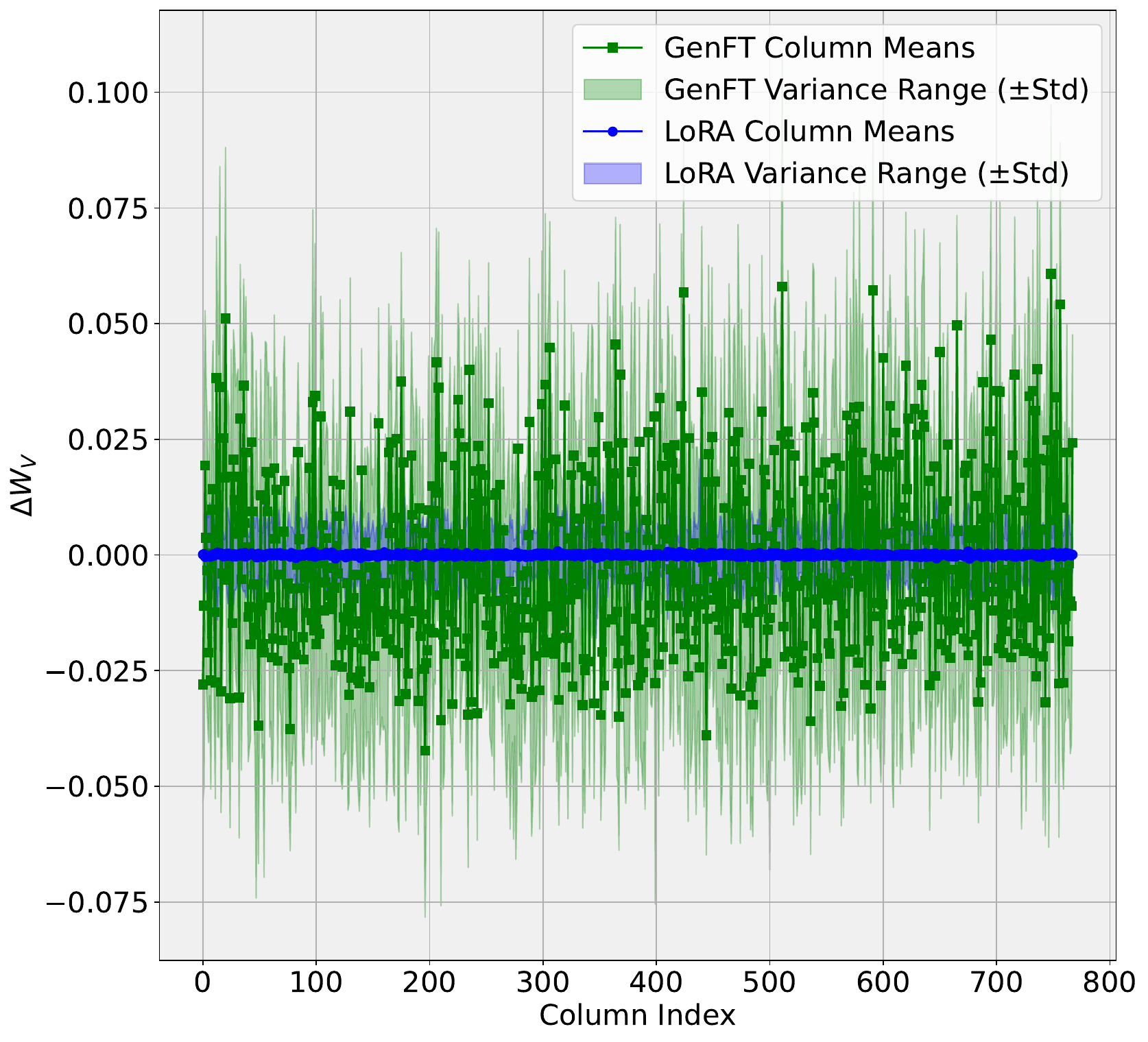}
        \caption{GenFT $\Delta W_V$ vs LoRA $\Delta W_V$}
    \end{subfigure}
    \caption{Column information comparison}
    \label{fig:genft_lora_column}
\end{figure}

We visualize row and column features on EuroSAT. Figures~\ref{fig:pretrained}--\ref{fig:genft} show the pretrained weights $W_Q$, $W_V$, and the adapted weights $W_0+\Delta W$ from LoRA and GenFT. LoRA makes only slight changes, suggesting a relatively uniform transformation, whereas GenFT captures more salient row and column patterns. Further comparisons in Figures~\ref{fig:genft_lora_row} and~\ref{fig:genft_lora_column} show that GenFT produces larger variations, indicating a weighted transformation that emphasizes important structures in $W_0$ and yields a more effective $\Delta W$.

\section{Initialization Method Analysis} 

We further analyze how different initialization methods affect the best-reported performance. We select three subsets from the three benchmarks and compare performance when varying the initialization scheme. The results are presented in Table~\ref{tab:intial}. Overall, the choice of initialization method has only a minor impact on the best performance, although an appropriate choice can still improve results.

\begin{table}[htbp]
\centering
\caption{Analysis results on different initial methods.}
\label{tab:intial}
\renewcommand\arraystretch{1.4}
\setlength{\tabcolsep}{2mm}
{
\begin{tabular}{c|c|ccc}
\hline
\rowcolor{gray!25}\textbf{Dataset} & \textbf{Type} & \textbf{A} & \textbf{B} & \textbf{Acc.}\\
\hline
\multirow{5}{*}{Clevr Count} & Reported & zero & normal & \textbf{83.62}\\
\cline{2-5}
& \multirow{4}{*}{Analyzed} & zero & zero & \underline{83.55}\\
&  & kaiming uniform & kaiming uniform & 82.77\\
&  & xavier uniform & xavier uniform & 82.35\\
&  & normal & normal & 82.97\\
\hline
\multirow{5}{*}{CUB} & Reported & normal & zero & 89.33\\
\cline{2-5}
& \multirow{4}{*}{Analyzed} & zero & zero & 89.54\\
&  & kaiming uniform & kaiming uniform & \underline{89.63}\\
&  & xavier uniform & xavier uniform & 89.54\\
&  & normal & normal & \textbf{89.78}\\
\hline
\multirow{5}{*}{MRPC} & Reported & zero & zero & \textbf{89.95}\\
\cline{2-5}
& \multirow{3}{*}{Analyzed} & kaiming uniform & kaiming uniform & 88.48\\
&  & xavier uniform & xavier uniform & 88.48\\
&  & normal & normal & \underline{89.46}\\
\hline
\end{tabular}}
\end{table}

\section{Experiments Results Details}
\label{app:experiments_details}
Here, Table~\ref{tab:detail_vtab1k} supplements the main comparison on VTAB-1K, presenting comprehensive results across all 19 datasets. Table~\ref{tab:detail_ablation_vtab1k} complements the ablation study on VTAB-1K, fully displaying ablation results for all 19 datasets. Table~\ref{tab:detail_ablation_fgvc} extends the FGVC datasets, providing complete ablation results.

\begin{table*}[ht!]
\centering
\caption{Benchmark results on VTAB-1K with ViT-B/16 models pre-trained on ImageNet-21K.}
\label{tab:detail_vtab1k}
\renewcommand\arraystretch{1.2}
\resizebox{\linewidth}{!}{
\begin{tabular}{c|c|ccccccc|cccc|cccccccc|c}
\hline
\rowcolor{gray!25}& & \multicolumn{7}{c|}{Natural} & \multicolumn{4}{c|}{Specialized} & \multicolumn{8}{c|}{Structured} & \\ 
\hline
\rowcolor{gray!25}\diagbox{Method}{Dataset} & \rotatebox{90}{\# Params. (M)} & \rotatebox{90}{CIFAR-100} & \rotatebox{90}{Caltech101} & \rotatebox{90}{DTD} & \rotatebox{90}{Flowers102} & \rotatebox{90}{Pets} & \rotatebox{90}{SVHN} & \rotatebox{90}{Sun397} & \rotatebox{90}{Patch Camelyon} & \rotatebox{90}{EuroSAT} & \rotatebox{90}{Resisc45} & \rotatebox{90}{Retinopathy} & \rotatebox{90}{Clevr/count} & \rotatebox{90}{Clevr/distance} & \rotatebox{90}{DMLab} & \rotatebox{90}{KITTI/distance} & \rotatebox{90}{dSprites/loc} & \rotatebox{90}{dSprites/ori} & \rotatebox{90}{SmallNORB/azi} & \rotatebox{90}{SmallNORB/ele} & \rotatebox{90}{Mean} \\ 
 \hline
Full FT~\cite{jia2022visual} & 85.88 & 68.9 & 87.7 & 64.3 & 97.2 & 86.9 & 37.4 & 38.8 & 79.7 & 95.7 & 84.2 & 73.9 & 56.3 & 58.6 & 41.7 & 65.5 & 57.5 & 46.7 & 25.7 & 29.1 & 65.57 \\
Linear~\cite{jia2022visual} & 0.00 & 63.4 & 85.0 & 63.2 & 97.0 & 86.3 & 36.6 & 51.0 & 78.5 & 87.5 & 68.6 & 74.0 & 34.3 & 30.6 & 33.2 & 55.4 & 12.5 &20.0 & 9.6 & 19.2 & 52.94 \\
Adapter~\cite{houlsby2019parameter} & 0.16 & 69.2 & 90.1 & 68.0 & 98.8 & 89.9 & 82.8 & 54.3 & 84.0 & 94.9 & 81.9 & 75.5 & 80.9 & 65.3 & 48.6 & 78.3 & 74.8 & 48.5 & 29.9 & 41.6 & 71.44 \\
VPT-Deep~\cite{jia2022visual} & 0.56 & 78.8 & 90.8 & 65.8 & 98.0 & 88.3 & 78.1 & 49.6 & 81.8 & 96.1 & 83.4 & 68.4 & 68.5 & 60.0 & 46.5 & 72.8 & 73.6 & 47.9 & 32.9 & 37.8 & 69.43 \\
LoRA~\cite{hu2022lora} & 0.29 & 67.1 & 91.4 & 69.4 & 98.8 & 90.4 & 85.3 & 54.0 & 84.9 & 95.3 & 84.4 & 73.6 & 82.9 & 69.2 & 49.8 & 78.5 & 75.7 & 47.1 & 31.0 & 44.0 & 72.25 \\ 
AdaptFormer~\cite{chen2022adaptformer} & 0.16 & 70.8 & 91.2 & 70.5 & 99.1 & 90.9 & 86.6 & 54.8 & 83.0 & 95.8 & 84.4 & 76.3 & 81.9 & 64.3 & 49.3 & 80.3 & 76.3 & 45.7 & 31.7 & 41.1 & 72.32 \\
BitFit~\cite{zaken2022bitfit} & 0.10 & 72.8 & 87.0 & 59.2 & 97.5 & 85.3 & 59.9 & 51.4 & 78.7 & 91.6 & 72.9 & 69.8 & 61.5 & 55.6 & 32.4 & 55.9 & 66.6 & 40.0 & 15.7 & 25.1 & 62.05 \\
SPT-LoRA~\cite{he2023sensitivity} & 0.43 & 73.5 & 93.3 & 72.5 & 99.3 & 91.5 & 87.9 & 55.5 & 85.7 & 96.2 & 75.9 & 85.9 & 84.4 & 67.6 & 52.5 & 82.0 & 81.0 & 51.1 & 30.2 & 41.3 & 74.07 \\
E$^2$VPT~\cite{he2023sensitivity} & 0.28 & 78.6 & 89.4 & 67.8 & 98.2 & 88.5 & 85.3 & 52.3 & 82.5 & 96.8 & 84.8 & 73.6 & 71.7 & 61.2 & 47.9 & 75.8 & 80.8 & 48.1 & 31.7 & 41.9 & 71.42 \\
SPT-Deep~\cite{wang2024revisiting} & 0.22 & 79.3 & 92.6 & 73.2 & 99.5 & 91.0 & 89.1 & 51.2 & 85.4 & 96.8 & 84.9 & 74.8 & 70.3 & 64.8 & 54.2 & 75.2 & 79.3 & 49.5 & 36.5 & 41.5 & 73.11 \\
SA$^2$VP~\cite{pei2024sa2vp} & 0.68 & 73.0 & 91.9 & 70.5 & 99.1 & 90.8 & 84.7 & 56.8 & 86.0 & 95.9 & 85.8 & 75.2 & 76.6 & 61.8 & 50.8 & 79.9 & 84.5 & 52.8 & 34.7 & 45.3 & 73.48\\
VFPT~\cite{zeng2024visual} & 0.45 & 80.7 & 91.4 & 69.4 & 99.3 & 90.3 & 85.6 & 52.7 & 83.5 & 96.5 & 84.4 & 75.4 & 75.8 & 63.2 & 48.3 & 79.3 & 81.5 & 56.0 & 34.1 & 43.4 & 73.20\\
FPET$_{\text{LoRA}}$~\cite{kim2025faster} & 0.30 & 70.1 & 92.7 & 69.4 & 99.1 & 90.8 & 85.4 & 55.6 & 87.2 & 94.6 & 82.5 & 74.1 & 83.0 & 63.4 & 50.6 & 81.6 & 84.7 & 51.5 & 34.3 & 43.3 & 73.36\\
FPET$_{\text{AdaptFormer}}$~\cite{kim2025faster} & 0.17 & 71.3 & 93.5 & 69.9 & 99.3 & 90.7 & 87.0 & 54.7 & 87.5 & 95.1 & 84.5 & 76.2 & 83.6 & 63.1 & 52.2 & 81.3 & 87.1 & 54.1 & 33.5 & 40.2 & 73.75\\
VAPT~\cite{le2025adaptive} & 0.19 & 80.8 & 91.9 & 69.7 & 98.8 & 89.2 & 86.7 & 52.9 & 84.4 &  96.5 & 85.1 & 74.5 & 74.8 & 63.6 & 50.0 & 77.2 & 86.1 & 48.3 & 33.8 & 40.9 & 72.91\\
Fact~\cite{jie2023fact} & 0.01 & 70.3 & 88.7 & 69.8 & 99.0 & 90.4 & 84.2 & 53.5 & 82.8 & 95.6 & 82.8 & 75.7 & 81.1 & 68.0 & 48.0 & 80.5 & 74.6 & 44.0 & 29.2 & 41.1 & 74.00\\
SSF+GIST~\cite{ruan2024gist} & 0.24 & 74.2 & 93.1 & 74.4 & 99.5 & 91.8 & 91.2 & 53.7 & 87.5 & 96.1 & 87.3 & 76.2 & 79.1 & 61.6 & 54.5 & 81.2 & 81.7 & 53.9 & 30.9 & 38.2 & 74.01\\
\hline
\rowcolor{gray!25}GenFT  & 0.27 & 71.4 & 92.9 & 72.4 & 99.0 & 91.4 & 90.9 & 56.2 & 87.1 & 96.0 & 87.4 & 76.4 & 83.6 & 64.7 & 53.9 & 81.6 & 82.1 & 49.9 & 35.3 & 43.3 & 74.50 \\
\hline
\end{tabular}
}
\end{table*}

\begin{table*}[ht!]
\centering
\caption{Ablation results on VTAB-1K with ViT-B/16 models pre-trained on ImageNet-21K.}
\scriptsize 
\label{tab:detail_ablation_vtab1k}
\renewcommand\arraystretch{1.4}
\resizebox{\linewidth}{!}{
\begin{tabular}{c|c|ccccccc|cccc|cccccccc|c}
\hline
\rowcolor{gray!25}& & \multicolumn{7}{c|}{Natural} & \multicolumn{4}{c|}{Specialized} & \multicolumn{8}{c|}{Structured} & \\ 
\hline
\rowcolor{gray!25}\diagbox{Method}{Dataset} & \rotatebox{90}{\# Params. (M)} & \rotatebox{90}{CIFAR-100} & \rotatebox{90}{Caltech101} & \rotatebox{90}{DTD} & \rotatebox{90}{Flowers102} & \rotatebox{90}{Pets} & \rotatebox{90}{SVHN} & \rotatebox{90}{Sun397} & \rotatebox{90}{Patch Camelyon} & \rotatebox{90}{EuroSAT} & \rotatebox{90}{Resisc45} & \rotatebox{90}{Retinopathy} & \rotatebox{90}{Clevr/count} & \rotatebox{90}{Clevr/distance} & \rotatebox{90}{DMLab} & \rotatebox{90}{KITTI/distance} & \rotatebox{90}{dSprites/loc} & \rotatebox{90}{dSprites/ori} & \rotatebox{90}{SmallNORB/azi} & \rotatebox{90}{SmallNORB/ele} & \rotatebox{90}{Mean} \\ 
 \hline
w/o shared & 0.09 & 66.6 & 89.7 & 72.6 & 98.5 & 90.0 & 71.3 & 55.9 & 82.8 & 95.2 & 84.4 & 75.2 & 61.4 & 36.5 & 49.2 & 74.8 & 18.1 & 23.1 & 15.2 & 39.9 & 63.18 \\
w/o specific & 0.23 & 67.6 & 92.5 & 71.1 & 98.6 & 91.0 & 90.9 & 56.0 & 87.1 & 96.0 & 87.4 & 75.0 & 83.0 & 63.7 & 54.4 & 81.3 & 6.8 & 30.8 & 28.7 & 41.6 & 68.61 \\
w/o row & 0.27 & 66.0 & 92.4 & 72.7 & 98.9 & 91.7 & 90.7 & 56.3 & 87.0 & 95.4 & 87.1 & 75.5 & 82.5 & 64.7 & 53.3 & 81.9 & 45.1 & 30.3 & 28.3 & 40.8 & 70.56 \\
w/o column & 0.27 & 67.1 & 92.3 & 72.7 & 99.0 & 91.1 & 90.3 & 56.9 & 85.7 & 95.3 & 87.3 & 76.4 & 82.6 & 63.7 & 53.8 & 79.5 & 43.1 & 30.6 & 26.9 & 41.8 & 70.32 \\
\hline
\rowcolor{gray!25}GenFT  & 0.27 & 71.4 & 92.9 & 72.4 & 99.0 & 91.4 & 90.9 & 56.2 & 87.1 & 96.0 & 87.4 & 76.4 & 83.6 & 64.7 & 53.9 & 81.6 & 82.1 & 49.9 & 35.3 & 43.3 & 74.50 \\
\hline
\end{tabular}
}
\end{table*}

\begin{table*}[ht!]
\centering
\caption{Ablation results on FGVC with ViT-B/16 models pre-trained on ImageNet-21K.}
\small
\label{tab:detail_ablation_fgvc}
\resizebox{\linewidth}{!}{
\begin{tabular}{c|ccccccc}
\hline
\rowcolor{gray!25}\diagbox{Method}{Dataset} & \makecell{Params. \\(M)} & \makecell{CUB200 \\(Acc.)} & \makecell{NABirds \\(Acc.)} & \makecell{Flowers \\(Acc.)} & \makecell{Dogs \\(Acc.)} & \makecell{Cars \\(Acc.)} & Avg.\\
\hline
w/o shared & 0.06 & 87.7 & 84.6 & 99.2 & 90.0 & 60.0 & 84.30 \\
w/o specific & 0.27 & 89.4 & 85.4 & 99.2 & 91.2 & 86.2 & 90.28 \\
w/o row & 0.27 & 89.1 & 85.2 & 99.2 & 91.1 & 85.2 & 89.96 \\
w/o column & 0.27 & 89.2 & 85.3 & 99.2 & 90.9 & 84.2 & 89.76 \\
\hline
\rowcolor{gray!25}GenFT & 0.29 & 89.3 & 85.6 & 99.3 & 91.2 & 86.2 & 90.38 \\
\hline
\end{tabular}
}
\end{table*}

\bibliographystyle{splncs04}
\bibliography{ref}

\end{document}